\documentclass[10pt,twocolumn,letterpaper]{article}

\usepackage[pagenumbers]{cvpr} 
\usepackage{placeins}
\usepackage{float}
\usepackage{multirow}
\usepackage{pifont}
\usepackage{makecell}
\usepackage{csquotes}
\definecolor{my_green}{RGB}{51,102,0}
\definecolor{my_red}{RGB}{204, 0, 0}
\newcommand{\cmark}{\textcolor{my_green}{\ding{51}}} 
\newcommand{\xmark}{\textcolor{my_red}{\ding{55}}} 
\usepackage{xargs}
\usepackage[colorinlistoftodos,prependcaption,textsize=normalsize]{todonotes}
\newcommandx{\xiangyu}[2][1=]{\todo[linecolor=red,backgroundcolor=red!25,bordercolor=red,#1]{#2}}

\usepackage[dvipsnames]{xcolor}

\usepackage{marvosym}  
\usepackage{graphicx}
\usepackage{caption}
\usepackage{stfloats} 
\usepackage{cuted}
\usepackage{multirow}
\usepackage{tabularx}

\usepackage{tocloft}
\usepackage{etoc}

\usepackage{titletoc}

\newcommand\DoToC{%
  \startcontents
  \printcontents{}{1}{\noindent \textbf{\Large{Contents in Appendix}}\vskip3pt\vskip5pt}
  \vskip3pt\vskip5pt
}

\newcommand{\listappendixname}{List of Appendices}
\newlistof{appendices}{apc}{\listappendixname}
\newlistof{appfigures}{afg}{List of Case Study Figures}

\usepackage[normalem]{ulem}
\useunder{\uline}{\ul}{}

\extrafloats{100}

\usepackage{multicol}
\usepackage{aliascnt}
\usepackage{adjustbox}

\marginparwidth=\dimexpr \marginparwidth + 1.5cm\relax
\definecolor{cvprblue}{rgb}{0.21,0.49,0.74}
\usepackage[pagebackref,breaklinks,colorlinks,allcolors=cvprblue]{hyperref}

\def\paperID{10299} 
\def\confName{CVPR}
\def\confYear{2025}

\title{AV-Odyssey Bench: Can Your Multimodal LLMs Really Understand Audio-Visual Information?}

\author{
    Kaixiong Gong$^1$\thanks{Equal Contribution},\; 
    Kaituo Feng$^1$\footnotemark[1],\;
    Bohao Li$^2$\footnotemark[1] \thanks{Project Leader},\;
    Yibing Wang,
    Mofan Cheng, \\
    Shijia Yang$^3$,\;
    Jiaming Han$^1$,\;
    Benyou Wang$^2$\thanks{Corresponding Authors: \small \texttt{wangbenyou@cuhk.edu.cn} and \texttt{xyyue@ie.cuhk.edu.hk}},\;
    Yutong Bai$^4$,\;
    Zhuoran Yang$^5$,\;
    Xiangyu Yue$^1$\footnotemark[3] \\[2mm]
    $^1$CUHK MMLab,
    $^2$CUHK (SZ), \\ 
    $^3$Stanford University, $^4$UC Berkeley, $^5$Yale University\\[2mm]
    \url{https://av-odyssey.github.io/}
}

\begin{document}
\maketitle
\begin{abstract}

Recently, multimodal large language models (MLLMs), such as GPT-4o, Gemini 1.5 Pro, and Reka Core, have expanded their capabilities to include vision and audio modalities. While these models demonstrate impressive performance across a wide range of audio-visual applications, our proposed DeafTest reveals that MLLMs often struggle with simple tasks humans find trivial: 1) determining which of two sounds is louder, and 2) determining which of two sounds has a higher pitch. 
Motivated by these observations, we introduce AV-Odyssey Bench, a comprehensive audio-visual benchmark designed to assess whether those MLLMs can truly understand the audio-visual information. This benchmark encompasses 4,555 carefully crafted problems, each incorporating text, visual, and audio components. To successfully infer answers, models must effectively leverage clues from both visual and audio inputs.
To ensure precise and objective evaluation of MLLM responses, we have structured the questions as multiple-choice, eliminating the need for human evaluation or LLM-assisted assessment. We benchmark a series of closed-source and open-source models and summarize the observations. By revealing the limitations of current models, we aim to provide useful insight for future dataset collection and model development.

\end{abstract}    
\section{Introduction}
\label{sec:intro}

Multimodal Large Language Models have evolved progressively, beginning with vision language models. Vision Language Models (VLMs), exemplified by GPT-4V(ision) \cite{gpt4}, have endowed language models with visual perception, enabling them to tackle a wide array of vision-language tasks \cite{yang2023set}. These models demonstrate remarkable capabilities, including counting objects in images~\citep{xu2023zerocounting}, performing numerical calculations on tabular data~\cite{yang2023dawn}, and solving geometric problems with provided figures \cite{mathverse}. Building upon this foundation, Multimodal Large Language Models (MLLMs)\footnote{In this paper, MLLMs only refer to the audio-vision LLMs. We use VLM to refer to the vision-language LLMs.} have further expanded their capabilities by incorporating audio modality, \eg, GPT-4o \cite{gpt4-o} and Gemini 1.5 \cite{gemini1.5}. These advancements push the boundaries of multimodal reasoning, particularly in areas including automatic speech recognition (ASR) \cite{gpt4-o}, automatic speech translation (AST) \cite{gemini1.5}, audio-visual captioning~\cite{onellm,anygpt}, and general audio-visual processing~\cite{onellm}.

\begin{figure}
\centering
\includegraphics[width=\linewidth]{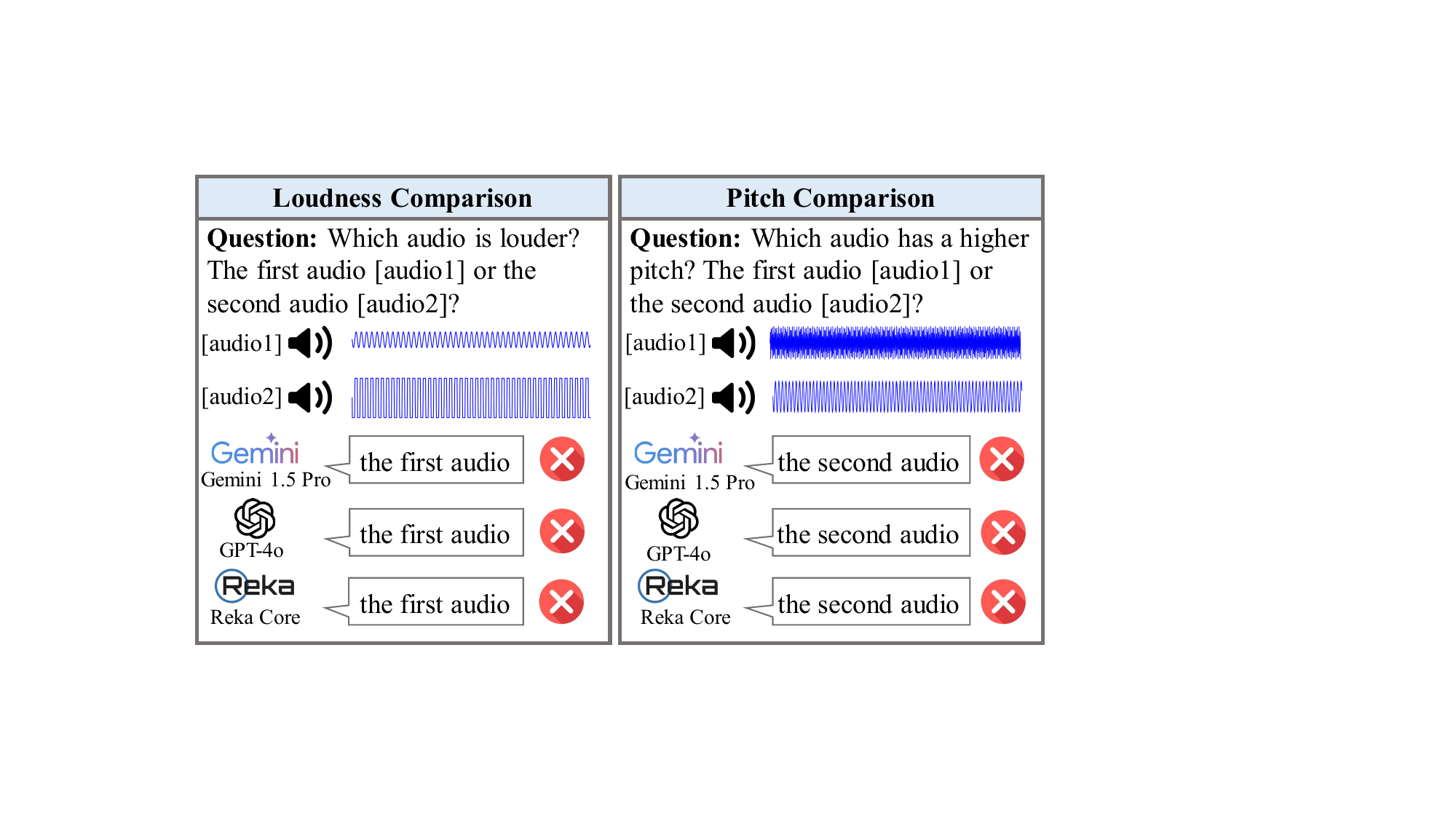}
\vspace{-0.2in}
\caption{Illustration of two out of four DeafTest tasks. Loudness comparison is used to determine the louder sound of two given sounds. Pitch comparison is to determine which sound has the higher pitch.}
\label{fig:deaftest}
\vspace{-0.15in}
\end{figure}

\begin{figure*}
\centering
\includegraphics[width=\textwidth]{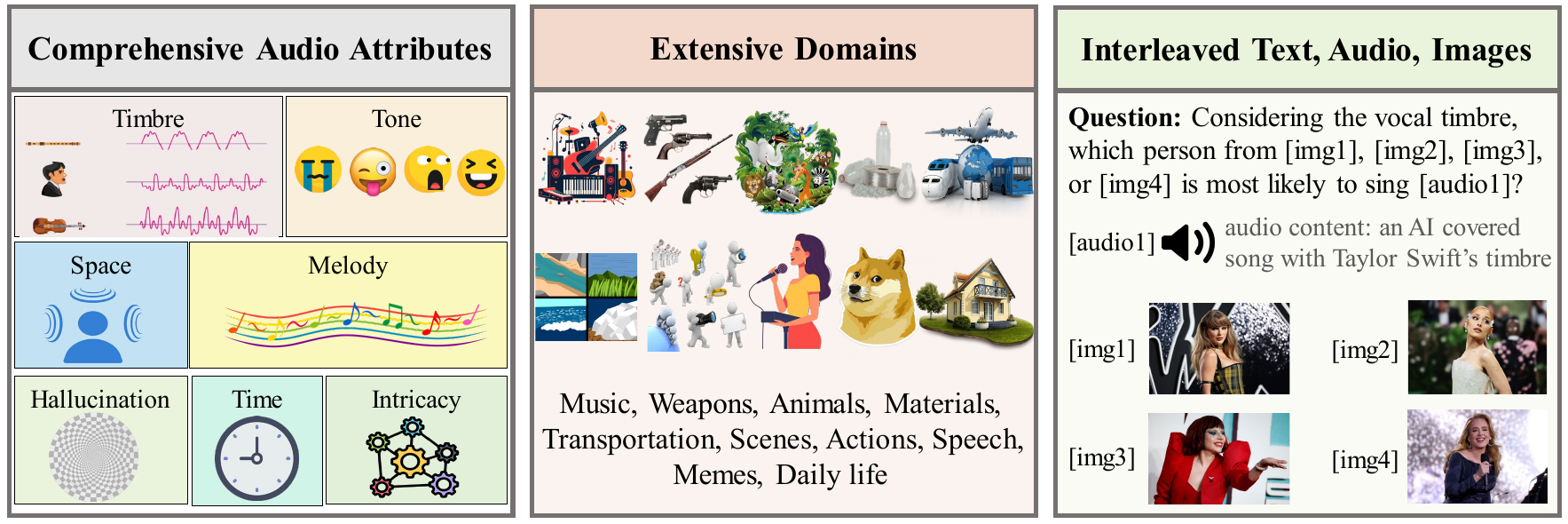}
\vspace{-0.2in}
\caption{Overview of AV-Odyssey Benchmark. AV-Odyssey Bench demonstrates three major features: 1. Comprehensive Audio Attributes; 2. Extensive Domains; 3. Interleaved Text, Audio, and Images.}
\label{fig:overview}
\vspace{-0.1in}
\end{figure*}

Most previous multimodal benchmarks focus on visual problem-solving~\cite{liu2023mmbench,yu2023mmvet, li2023seed}, \eg, general vision comprehension~\cite{seedbench2,liu2023mmbench,fu2023mme} and multimodal mathematical reasoning~\cite{geoqa1,geoqa2,unigeo,mathverse,lu2023mathvista,lu2021inter}. To evaluate audio-visual comprehension, dedicated benchmarks such as AVQA~\cite{avqa} and MusicAVQA~\cite{musicavqa} have been introduced, assessing models in several audio-visual tasks~\cite{avqa,musicavqa,li2024omnibench,panoavqa,savebench}.


However, we primarily identify two issues of previous audio-visual benchmarks: 1) They \textbf{overlook assessing the basic \textit{listening} ability (no reasoning)} of MLLMs.  2) They are \textbf{limited in evaluation dimensions} (\eg, restricted audio attributes and narrow application domains), as illustrated in Table~\ref{differ}. This leads to an insufficient evaluation of the audio-visual information understanding capability of MLLMs. These findings raise the central question we aim to investigate in this paper: $\bigstar$ \textbf{Can Multimodal LLMs Truly Understand Audio-Visual Information?} To investigate this, we propose DeafTest to assess the basic \textit{listening} ability of MLLMs, along with the comprehensive AV-Odyssey Benchmark to further assess their audio-visual understanding capabilities.



\textbf{DeafTest}, a suite of four low-level auditory tasks (examples are illustrated in Figure~\ref{fig:deaftest}), is introduced to examine the fundamental ability of MLLMs to \textit{listen}, inspired by the Schwabach test in audiology~\cite{huizing1975early} and BlindTest~\cite{blindtest} in vision domain. We evaluate Gemini 1.5~\cite{gemini1.5}, Reka~\cite{team2024reka}, and GPT-4o~\cite{gpt4-o} on our four simple tasks that only involve basic elements of sound (\eg, loudness, pitch, and duration) as shown in Table~\ref{table:api_model_basic_performace}. The key findings are:
\begin{enumerate}
    \item Despite their ability to recognize complex speech content, MLLMs do not perform as well as expected on sound counting tasks. The best-performing model, Gemini 1.5 Pro, achieves only 81\%, while humans can easily score 100\%. The sounds in these tasks are monotonous and are clearly separated by silent intervals within the audio clip.
    \item MLLMs appear to be insensitive to sound volume. When two sounds, one louder than the other, are fed into models, they often fail to distinguish the louder one. All models perform below 65\%, significantly lower than the expected 100\%.
    \item MLLMs also struggle to discriminate the higher pitch between two given sounds. None of the models can score above 60\%.
    \item The duration comparison task presents models with two sounds and asks them to determine which has the longer duration. Two Gemini models 
    perform significantly better than the others that are merely on par with the random baseline. 
\end{enumerate} The results suggest that \textbf{while MLLMs excel at some high-level multimodal reasoning tasks, they still have limitations in basic \textit{listening} abilities, which may further hinder their integration of audio-visual information }(hypothesis $\blacktriangle$\label{hypothesis}). DeafTest allows us to critically assess whether MLLMs truly understand audio-visual information or if their apparent capabilities are limited to surface-level pattern recognition. By doing so, we aim to contribute to a deeper understanding of multimodal comprehension in current MLLMs and identify areas for future improvement—issues that have been overlooked by previous audio-visual benchmarks~\cite{avqa,musicavqa,li2024omnibench}.


\begin{table}[t]
    \begin{center}
    \setlength\tabcolsep{1pt}
    \footnotesize
    \caption{Results on four basic auditory tasks (DeafTest). The questions are designed as two-choice questions. The random baseline performance is 50\%. }
    \vspace{-0.1in}
    \begin{tabular}{lcccc}
\toprule
    \multirow{1}{*}{Method} & \makecell{Sound \\ Counting} & \makecell{Loudness \\ Comparison} & \makecell{Pitch \\ Comparison} & \makecell{Duration \\ Comparison} \\
    \cmidrule(lr){1-5}
    Random & 50.0 & 50.0 & 50.0 & 50.0 \\
    Gemini 1.5 Flash ~\citep{gemini1.5} & 55.0 & 62.0 & 54.0 & 89.0\\
    Gemini 1.5 Flash-8B~\citep{gemini1.5} & 49.0 & 55.0 & 51.0 & 51.0\\
    Gemini 1.5 Pro~\citep{gemini1.5} & 81.0 & 60.0 & 52.0 & 84.0\\
    Reka Core~\citep{team2024reka} & 54.0 & 43.0 & 42.0 & 40.0 \\
    Reka Flash~\citep{team2024reka} & 48.0 & 58.0 & 51.0 & 44.0 \\
    Reka Edge~\citep{team2024reka} & 47.0 & 56.0 & 50.0 & 44.0\\
    GPT-4o audio-preview~\citep{gpt4-o} & 50.0 & 58.0 & 58.0 & 57.0\\
    \bottomrule
    \label{table:api_model_basic_performace}
    \end{tabular}
    \end{center}
    \vspace{-0.365in}
\end{table}

\begin{table*}[t]
    \centering
    \caption{Comparisons between MLLM benchmarks / datasets.}
    \vspace{-0.1in}
    \small
    \setlength{\tabcolsep}{0.5mm}
      \resizebox{\textwidth}{!}{
  \begin{tabular}{@{}cccccccccccccc@{}}
  \toprule
  \multirow{2}{*}{Benchmark / Dataset} & \multirow{2}{*}{Modality} & \multirow{2}{*}{Questions} & \multirow{2}{*}{Answer Type} & \multirow{2}{*}{\begin{tabular}[c]{@{}c@{}}Customized\\ Question\end{tabular}} & \multicolumn{7}{c}{Audio Attributes}                              & \multirow{2}{*}{\begin{tabular}[c]{@{}c@{}}Multiple\\ Domains\end{tabular}} & \multirow{2}{*}{Interleaved} \\ \cmidrule(lr){6-12}
                             &                 &          &                            &                              & Timbre & Tone & Melody & Space & Time & Hallucination & Intricacy &                                                                             &                              \\ \midrule
  MME Bench~\cite{fu2023mme}              & Image         & 2194              & Y/N  & \cmark          & -        & -     & -       & -      & -     & -              & -                 & \cmark        & \xmark   \\
  MMBench~\cite{liu2023mmbench}         & Image(s)         & 2974              &  A/B/C/D  & \cmark         & -        & -     & -       & -      & -     & -              & -        & \cmark       & \xmark    \\
  SEED-Bench-2~\cite{seedbench2}     & Image(s) \& Video         & 24371             & A/B/C/D   & \cmark        & -        & -     & -       & -      & -     & -              & -             & \cmark     &  \cmark     \\ \midrule
  AVQA Dataset~\cite{avqa}             & Video \& Audio         & 57335             & A/B/C/D    & \cmark        & \cmark       & \xmark      & \xmark        & \xmark       & \cmark     & \xmark              & \cmark                 & \cmark        & \xmark    \\
  Pano-AVQA Dataset~\cite{panoavqa}        &  Video \& Audio        & 51700             & defined words \& bbox  & \cmark          &  \cmark      & \cmark     & \xmark       & \cmark      & \xmark     & \xmark              &  \cmark                &  \cmark    &   \xmark    \\
  Music-AVQA Dataset~\cite{musicavqa}       &  Video \& Audio        & 45867             & defined words  & \cmark          &  \cmark      & \xmark     & \cmark       & \cmark      & \cmark      & \cmark              &  \cmark                &  \xmark       & \xmark   \\ \midrule
  SAVE Bench~\cite{savebench}             & Image \& Video \& Audio         & 4350             & free-form & \xmark          & \cmark        & \xmark      & \xmark        &  \cmark      & \xmark      & \xmark               &   \cmark                & \cmark       & \xmark    \\
  OmniBench~\cite{li2024omnibench}       & Image \& Audio         & 1142             & A/B/C/D  & \cmark           & \cmark       & \xmark     & \xmark       & \xmark      & \cmark     & \xmark              &  \xmark                & \cmark          & \xmark \\ \midrule
  AV-Odyssey Bench (ours) & Image(s) \& Video \& Audio(s)         & 4555             & A/B/C/D   & \cmark    & \cmark       & \cmark     & \cmark       & \cmark      & \cmark     &  \cmark             &   \cmark               & \cmark      & \cmark     \\ \bottomrule
  \end{tabular}
    }
    \label{differ}
    \vspace{-0.1in}
  \end{table*}

To further investigate question $\bigstar$, we introduce \textbf{AV-Odyssey}, a comprehensive audio-visual benchmark designed to challenge MLLMs by requiring them to leverage information from all input modalities to derive accurate answers. This meticulously crafted dataset encompasses 4,555 carefully selected questions spanning 26 distinct tasks, with each question strategically constructed to include three critical components: text, image/video, and audio clip.


To ensure the benchmark's rigor, we employ vision and audio language models to filter out questions that could be easily resolved by single-modal approaches. This guarantees that only complex, multi-dimensional questions remain, truly testing the models' ability to integrate information across modalities. AV-Odyssey is designed with a broad scope, covering a wide range of sound attributes, including timbre, tone, spatial characteristics, and temporal dynamics, while simultaneously exploring diverse application domains such as music, daily life, and transportation, as illustrated in Figure~\ref{fig:overview}. By structuring the evaluation as multiple-choice selections, we've streamlined the assessment process, eliminating the need for manual verification or LLM-assisted evaluation.

We conduct extensive experiments on closed-source model~\cite{gpt4-o,gemini1.5, team2024reka} and open-source models~\cite{onellm,unifedio,anygpt,wu2023nextgpt, su2023pandagpt, videollama2, fu2024vita} on the proposed AV-Odyssey Bench. Considering the results of DeafTest and AV-Odyssey, we have the following findings:
\begin{itemize}
    \item Overall, current MLLMs still fall short in processing complex audio-visual information integration tasks. 
    \item The audio-caption-vision training paradigm fails to effectively combine audio and visual modalities, limiting the integration of audio-visual information (see Sec.~\ref{sec:finding3}). 
    \item Through error analysis (see Sec.~\ref{sec:error_analysis}), we find out that the major cause of error in audio-visual inference is the perceptual error of audio input, which is in line with the hypothesis $\blacktriangle$.
\end{itemize}

In a nutshell, this paper explores \textbf{whether MLLMs really understand audio-visual information} from two aspects. To begin with, we propose \textbf{DeafTest} to evaluate MLLMs and find out that these MLLMs still have obvious limitations in basic \textit{listening} ability. This could lead to a bottleneck in subsequent audio-visual information integration for complex audio-visual reasoning, which is then validated by the results and analysis of our proposed \textbf{AV-Odyssey benchmark}.

\section{Related Work}
\label{sec:related work}

\textbf{Multimodal Large Language Models}. 
Large language models (LLMs) have demonstrated remarkable performance across diverse textual domains~\cite{gpt4,gpt1,gpt2, llama,bi2024deepseek}. The success of these models has catalyzed significant advancements in vision language models and multimodal large language models. Inspired by the textual prowess of LLMs, vision language models have emerged to extend computational capabilities into visual comprehension. These models enable LLMs to perform sophisticated visual tasks, including visual question answering~\cite{liu2023llava1.5, li2023blip2, zhu2023minigpt4, liu2023visual_llava, ye2023mplugowl, dai2023instructblip, bai2023qwen, zhang2023internlm}, visual grounding~\citep{peng2023kosmos, wang2023cogvlm, chen2023minigpt, chen2023shikra}, document understanding~\citep{ye2023mplugowl, hu2024mplug, zhang2023llavar, lv2023kosmos}, long video understanding~\citep{liu2024kangaroo, shen2024longvu, zhang2024longva, li2023llamavid, ren2024timechat}. Building upon vision-language achievements, researchers have further expanded multimodal horizons by integrating the audio modality~\citep{onellm,unifedio,anygpt,wu2023nextgpt,su2023pandagpt,fu2024vita,videollama2}. These advanced models now accommodate audio inputs, further expanding the landscape of multimodal artificial intelligence.

\textbf{Benchmarking Multimodal Large Language Models}.
The rapid development of vision language models has been accompanied by the emergence of specialized benchmarks to assess their performance across various domains~\citep{mmmu,fu2023mme,seedbench2,geoqa1,lu2023mathvista}. A significant subset of these benchmarks focuses on vision comprehension~\citep{mmmu,fu2023mme,seedbench2, li2024seed2plus} and mathematical reasoning capabilities~\citep{geoqa1,geoqa2,lu2023mathvista,mathverse,mmmu,seo2015solving}.
However, current audio-visual benchmarks~\citep{avqa,musicavqa,panoavqa,savebench,li2024omnibench} face significant limitations in comprehensively assessing multimodal large language models (MLLMs). Firstly, they predominantly focus on high-level visual tasks and neglect to explore the basic auditory perception limitations. Secondly, they are limited in application domains. For example, Music-AVQA~\cite{musicavqa} limits evaluation to the music domain, and AVQA~\cite{avqa} primarily tests daily life applications. Thirdly, they do not comprehensively evaluate all attributes of the audio. In contrast, this paper introduces DeafTest tasks to evaluate fundamental capabilities and the AV-Odyssey benchmark, which spans a wide spectrum of audio attributes and application domains, enabling a comprehensive assessment of the audio-visual comprehension performance of MLLMs.

\section{Method}

\subsection{DeafTest Tasks}
Drawing inspiration from the Schwabach test~\cite{huizing1975early}, we introduce DeafTest, a suite of four simple auditory tasks that critically examine the fundamental audio perception capabilities of Multimodal Large Language Models (MLLMs). DeafTest includes the determination of the number of sounds, identification of the louder sound, recognition of the sound with a higher pitch, and detection of the sound with a longer duration. We hypothesize that MLLMs may not perform as well as expected on these basic tasks. This potential shortcoming arises from the training objectives of these models, which primarily focus on achieving high-level semantic alignment between different modalities. Consequently, this approach tends to overlook the effective utilization of low-level auditory information, which is crucial for accurately processing and understanding basic sound characteristics.


\textbf{1. Count the Number of Sounds}. 
Given that Multimodal LLMs achieve impressive performance on ASR (GPT-4o's 3\% word error rate on ASR Western Europe)~\citep{hello_gpt4o}, we expect that counting the number of sounds is not difficult for MLLMs. In this task, we give an audio clip that contains several sounds ranging from 3 to 8 and ask MLLMs for the number of sounds. In an audio clip, the sounds are monotonous and clearly separated by a silent clip. We formulate these queries as two-choice questions. The MLLMs only need to predict the correct option. The question number for this task, as well as for all remaining tasks in DeafTest, is set to 100.

\textbf{2. Discriminate the Louder Sound}. 
\label{sec:task2}
In this task, we test the basic ability of MLLMs to distinguish between the loudness of sounds. The goal of MLLMs is to discriminate which sound is louder out of two given audio clips. Specifically, the decibel for quieter audio ranges from 30 dB to 60 dB, while the decibel for louder audio ranges from 70 dB to 100 dB. We randomly sample decibels from these two ranges to create two audio clips. In addition, we randomly switch the input order of the two audio clips; that is, for some questions, the quiet audio comes first, and for the rest, the loud audio comes first. Similarly, the question format is also a two-choice question.

\textbf{3. Discriminate the Higher Pitch}. 
This task is similar to task 2 in ~\ref{sec:task2}. We also create two audio clips. The key difference between the two audios is pitch. Pitch is the basic element of sound, which is helpful in discriminating tone, emotion, environment, \textit{etc}. For the lower pitch audio, we randomly sample its pitch from 100Hz to 500Hz, while we randomly sample pitch from 1000Hz to 2000Hz for the higher pitch audio. We manually check these sounds to ensure that humans can easily discriminate between different pitches. In task 2, we randomly switch the input order of the two audio clips.

\textbf{4. Recognize the Duration of Sound}.
We also test MLLMs with the duration of sound. In this task, we simplify the question by giving two audio clips of different durations. We sample the duration from 1s to 3s for the short audio, while we sample from 4s to 6s for the long audio. Similar to task 2, we provide the MLLMs with two audio clips, asking them to identify the longer one.

The results on DeafTest are shown in Table~\ref{table:api_model_basic_performace}. Among the four tasks, sound counting and duration separation seem to be simpler than the other two for MLLMs, since Gemini 1.5 Pro achieves more than 80\% accuracy on the two tasks. Nonetheless, all the results are far behind the expected 100\%. Especially on loudness comparison and pitch comparison tasks, none of these MLLMs achieve a score over 65\%. All these results suggest that MLLMs fall short in basic \textit{listening} instinct, which might hinder further audio-visual information integration for solving sophisticated audio-visual comprehension tasks. To verify this, we further introduce a holistic AV-Odyssey benchmark to comprehensively evaluate the audio-visual performance of MLLMs as depicted in the following.

\begin{table}[]
  \centering
  \caption{Detailed statistics of AV-Odyssey Benchmark.}
  \vspace{-0.1in}
  \small
    \resizebox{\linewidth}{!}{%
  \begin{tabular}{@{}cc@{}}
\toprule
Statistics                                   & Number              \\ \midrule
Total Questions                              & 4555                \\
Total Tasks                                  & 26                  \\
Domains                                      & 10                  \\ \midrule
Questions with Multiple Images, Singe Audio  & 2610                \\
Questions with Single Image, Multiple Audios & 891                 \\
Questions with Singe Image, Singe Audio      & 434                 \\
Questions with Singe Video, Singe Audio      & 220                 \\
Questions with Single Video, Multiple Audios & 400                 \\ \midrule
Correct Option Distribution (A:B:C:D)        & 1167:1153:1119:1116 \\ \midrule
Average Audio Time                           & 16.32 seconds       \\
Average Image Resolution                     & 1267.72 $\times$ 891.40    \\
Average Video Resolution                     & 1678.69 $\times$ 948.56    \\
Average Video Time                           & 15.58 seconds       \\ \bottomrule
\end{tabular}
  }
  \label{tab:statistics}
  \vspace{-0.18in}
\end{table}

\begin{figure}
\centering
\includegraphics[width=\linewidth]{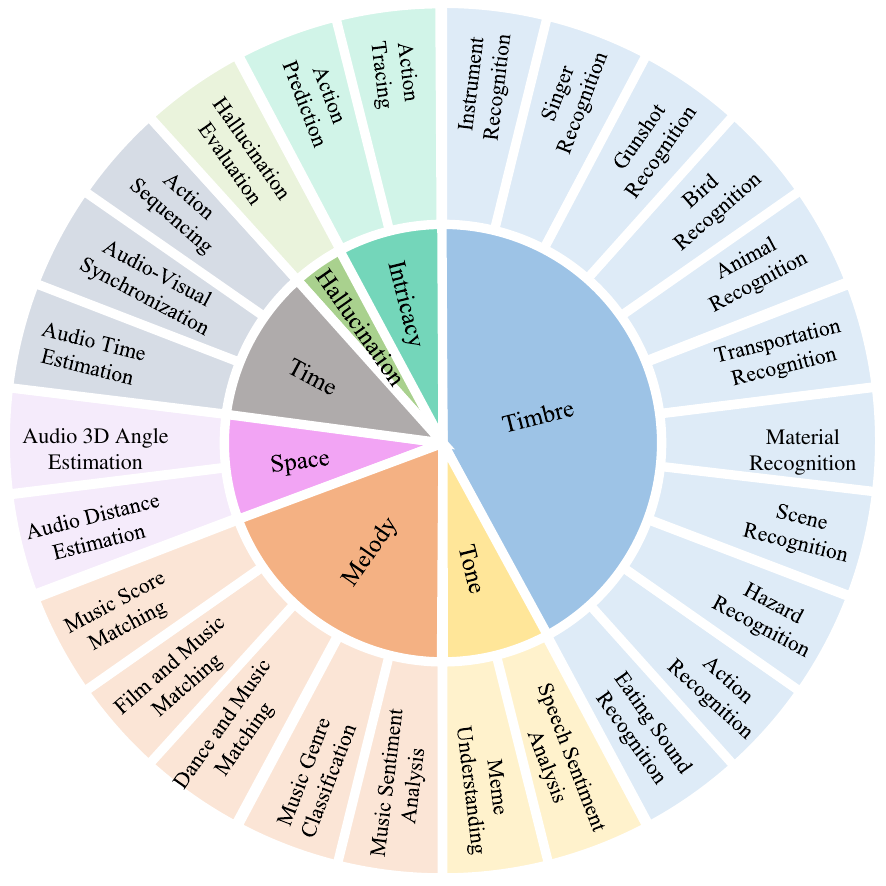}
\caption{Overview of 26 evaluation tasks of AV-Odyssey Benchmark. We mainly categorize these tasks with the sound attributed into 7 classes.}
\label{fig:task_ditribution}
\vspace{-0.22in}
\end{figure}

\begin{figure*}
\centering
\includegraphics[width=\textwidth]{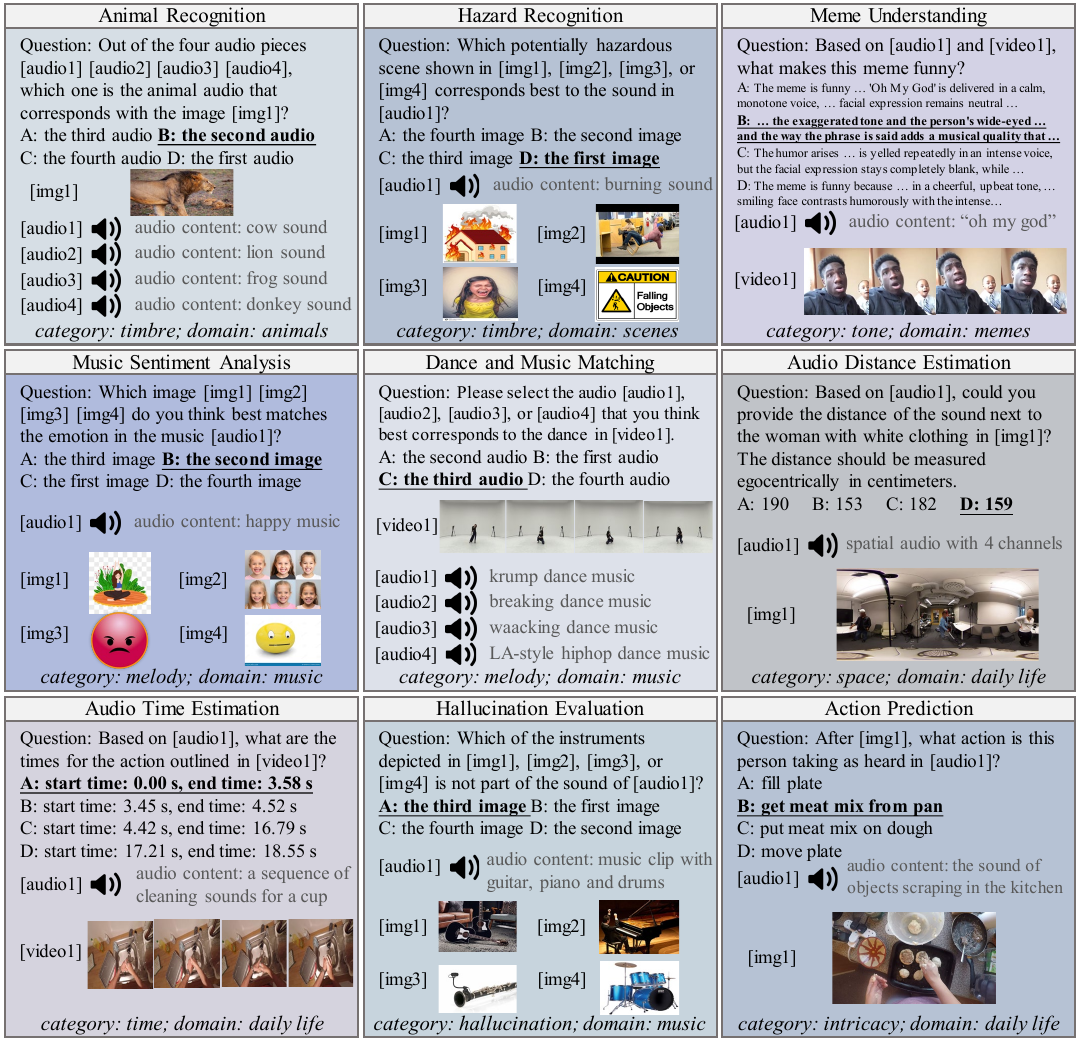}
\caption{Sampled examples from our AV-Odyssey Benchmark.}
\label{fig:demofig}
\vspace{-3mm}
\end{figure*}

\subsection{Overview of AV-Odyssey Bench}
Our AV-Odyssey Bench is a meticulously curated benchmark designed to comprehensively assess the audio-visual capabilities of MLLMs. 
To ensure a robust and unbiased assessment, all questions in AV-Odyssey are structured as multiple-choice, with four options per question, and options can be presented in various formats, including text, images, or audio clips.
To mitigate format-specific biases, we have curated five distinct multi-choice question types.  Additionally, all inputs, including text, image/video, and audio clips, are fed into MLLMs in an interleaved manner.

We compare our AV-Odyssey benchmark with previous MLLM benchmarks and datasets in Table~\ref{differ}. 
It can be found that previous works suffer from certain limitations, such as restricted audio attributes, which fail to capture the full spectrum of auditory complexity;  narrow domain focus, limiting the generalizability of findings; and the absence of interleaved settings, crucial for assessing real-world multimodal integration capabilities. 
For instance, Music-AVQA~\cite{musicavqa} limits audio and visual data to the music domain, while OmniBench~\cite{li2024omnibench} lacks multiple audio attributes, making it difficult to comprehensively assess the capabilities of MLLMs in audio-visual tasks. In contrast, our AV-Odyssey encompasses 26 tasks across 10 diverse domains and includes 7 audio attributes, with interleaved and customized questions. The detailed statistics are shown in Table~\ref{tab:statistics}.
This design enables an exhaustive evaluation of MLLMs, providing a nuanced and thorough assessment of their performance in complex, real-world audio-visual scenarios. 

Next, we will briefly introduce the task categories that span a broad spectrum of audio attributes, including Timbre, Tone, Melody, Spatial characteristics, Temporal dynamics, and Hallucination detection. The detailed task distribution and task examples are shown in Figure~\ref{fig:task_ditribution} and Figure~\ref{fig:demofig}, respectively.

\textbf{Timbre Tasks.} In order to test the concept of matching across vision and audio modalities, MLLMs are required to match audio-visual pairs (e.g., lion's roar sound with lion images) in timbre tasks. In addition, we have designed advanced tasks that demand internal expert-level knowledge learned from the large-scale pretraining data to solve, such as singer recognition and bird species identification. 

\textbf{Tone Tasks.} These tasks target evaluating MLLMs with speech sentiment analysis and meme understanding. For example, meme understanding requires MLLMs to infer humorous reasons simultaneously from the voice tone and visual context. 
 
\textbf{Melody Tasks.} For evaluating melody understanding abilities, we propose melody tasks. For example, the dance and music matching task requires the MLLM to understand the melody of the music and identify the one that aligns with the dance in a video.
 
\textbf{Space Tasks.} To test the spatial inference ability with audio and visual information, space tasks require MLLMs to infer the distance of a certain object producing a sound or to determine the 3D angle.
 
\textbf{Time Tasks.} These tasks test the cross-modal matching and temporal correlation abilities at the same time. For example, audio time estimation requires MLLMs to determine the start and end time of an action. 

\textbf{Hallucination Tasks}. Inspired by POPE~\cite{pope} that indicates severe object hallucination existing in vision language models, we designed this task to assess the hallucination issue in audio-visual reasoning.

\textbf{Intricacy Tasks.} These tasks challenge MLLMs to perform integrated analysis or reasoning through both visual and audio inputs, leveraging multiple attributes. For example, action prediction requires models to infer actions based on visual elements alongside various audio attributes, such as timbre and timing.
    
These diverse tasks provide a rigorous and multifaceted assessment of MLLMs' audio-visual information integration capabilities, systematically probing the depth, nuance, and complexity of cross-modal perception and reasoning.

\subsection{Data Curation Process}
\textbf{Data Collection}. AV-Odyssey Bench is an audio-visual benchmark to evaluate whether MLLMs truly have audio-visual reasoning capability. Since the audio is the newly added modality by these omni-modal models and there is already an array of visual benchmarks, we put our attention on the attributes of sound in the benchmark construction. We first go through all the attributes of sound, such as timbre, tone, time, space, \textit{etc}. Then, we start from a specific attribute of sound and span the domains to cover a wide range of application domains, such as music, daily life, and transportation. 
We primarily use two strategies to construct questions: 1) For most concept-matching questions (e.g., bird recognition, material recognition), we gather audio clips from public datasets and crawl corresponding visual data from the internet to automatically generate questions and options. Human experts conduct post-evaluation and filter out low-quality questions. 2) For other questions (e.g., meme understanding, audio 3D angle estimation), we manually collect audio and visual data from the internet or public datasets, relying on human experts to craft the questions and options. The datasets we used are listed as follows: ~\citep{task1data, kaggletask3data, kaggletask4data, kaggletask5data, erhan_akbal_task6data, sterling2018isnn_task7data, owens2016visually_task7data, heittola_task8_data, kay2017kinetics_task9_10_data, kaggletask11data, kaggle_TESS_task12data, kaggle_SAVEE_task12data, kaggle_Crema_task12data, kaggle_RAVDESS_task12data, kaggle_EMOTIFY_task14data, kaggle_GTZAN_task15data, aist-dance-task16data, defferrard2016fma_task17data, zhang2024gtsinger_task18data, shimada2024starss23_task19_20data, damen2018scaling_task21_23_25_26data, tian2018audio_task22data, ostermann2023aam_task24data}.

\begin{table*}[]
    \centering
    \caption{Evaluation results of various MLLMs in different parts of AV-Odyssey Bench. The highest performance is highlighted in bold, while the second highest is underlined. $\bar{T}$ is the averaged accuracy across corresponding dimensions, and $R_{\bar{T}}$ is the rank based on the the averaged accuracy. \enquote{All Avg.} represents the averaged accuracy over all questions in our AV-Odyssey Bench.}
    \label{tab:performance}
    \vspace{-0.1in}
    \vspace{3pt}
    \small
    \resizebox{\textwidth}{!}{
    \begin{tabular}{ccccccccccccccccccc}
         \toprule
         \multicolumn{2}{c}{\multirow{2}{*}{Model}} & \multirow{2}{*}{\makecell{LLM \\ Size}}& \multicolumn{2}{c}{Timbre} & \multicolumn{2}{c}{Tone} & \multicolumn{2}{c}{Melody} & \multicolumn{2}{c}{Space} & \multicolumn{2}{c}{Time}& \multicolumn{2}{c}{Hallucination}& \multicolumn{2}{c}{Intricacy}& \multicolumn{2}{c}{All Avg.}\\
         \cmidrule(lr){4-5}
         \cmidrule(lr){6-7}
         \cmidrule(lr){8-9}
         \cmidrule(lr){10-11}
         \cmidrule(lr){12-13}
         \cmidrule(lr){14-15}
         \cmidrule(lr){16-17}
         \cmidrule(lr){18-19}
         &&& $\bar{T}$ & $R_{\bar{T}}$ & $\bar{T}$ & $R_{\bar{T}}$ & $\bar{T}$ & $R_{\bar{T}}$ & $\bar{T}$ & $R_{\bar{T}}$ & $\bar{T}$ & $R_{\bar{T}}$ & $\bar{T}$ & $R_{\bar{T}}$ & $\bar{T}$ & $R_{\bar{T}}$ & $\bar{T}$ & $R_{\bar{T}}$\\
         \midrule
          \multicolumn{2}{c}{Random} & - & 25.0 & - & 25.0 & - & 25.0 & - & 25.0 & - & 25.0 & - & 25.0 & - & 25.0 & - & 25.0 & -\\
         \midrule
         \multirow{10}{*}{\rotatebox{90}{Open Source}} & Unified-IO-2 L~\citep{unifedio} & 1B & 23.8 & 16 & 24.1 & 11 & 28.8 & 6 & 15.0 & 18 & 26.8 & 9 & 30.0 & 5 & 30.4 & 11 & 26.0 & 16 \\
         & Unified-IO-2 XL~\citep{unifedio} & 3B & 24.3 & 12 & 23.2 & 13 & 27.8 & 7 & 22.5 & 14 & 25.3 & 16 & 31.5 & 2 & 34.8 & 4 & 26.3 & 12 \\
         & Unified-IO-2 XXL~\citep{unifedio} & 7B & 26.3 & 6 & 22.7 & 15 & 26.4 & 12 & 32.5 & 4 & 26.8 & 9 & 24.5 & 14 & 33.8 & 7 & 27.2 & 6 \\
         & OneLLM~\citep{onellm} & 7B & 25.0 & 10 & 25.5 & 6 & 21.5 & 18 & 37.5 & 2 & \bf{29.3} & 1 & 25.5 & 11 & \bf{38.4} & 1 & 27.4 & 5 \\
         & PandaGPT~\citep{su2023pandagpt} & 7B & 23.5 & 17 & 23.2 & 13 & 27.6 & 10 & \bf{45.0} & 1 & 23.8 & 18 & 28.0 & 10 & 23.9 & 17 & 26.7 & 10 \\
         & Video-llama~\citep{videollama} & 7B & 25.5 & 7 & 22.3 & 16 & 24.4 & 17 & 30.0 & 6 & 26.2 & 13 & 25.0 & 12 & 30.7 & 10 & 26.1 & 14 \\
         & VideoLLaMA2~\citep{videollama2} & 7B & 24.1 & 13 & 25.5 & 6 & 26.4 & 14 & 30.0 & 6 & 27.2 & 8 & \bf{33.0} & 1 & 34.5 & 5 & 26.8 & 9 \\
         & AnyGPT~\citep{anygpt} & 7B &24.6 & 11 & 25.0 & 8 & 26.4 & 15 & 27.5 & 11 & \underline{29.2} & 2 & 29.0 & 6 & 25.7 & 15 & 26.1 & 15 \\
         & NExT-GPT~\citep{wu2023nextgpt} & 7B &23.2 & 18 & 20.9 & 17 & 27.8 & 9 & 30.0 & 6 & 28.8 & 3 & 28.5 & 8 & 23.6 & 18 & 25.5 & 17 \\
         & VITA~\citep{fu2024vita} & $8\times \text{7B}$ & 24.1 & 14 & 26.4 & 5 & 27.8 & 7 & 22.5 & 14 & 26.3 & 12 & 31.0 & 4 & \underline{36.8} & 2 & 26.4 & 11\\
         \midrule
         \multirow{8}{*}{\rotatebox{90}{Closed Source}} & Gemini 1.5 Flash ~\citep{gemini1.5} & - & 27.2 & 4 & 25.0 & 8 & 28.8 & 5 & 30.0 & 6 & 25.3 & 16 & 28.5 & 8 & 31.2 & 9 & 27.8 & 4 \\
         & Gemini 1.5 Flash-8B~\citep{gemini1.5} & - & 25.1 & 9 & 24.5 & 10 & 28.9 & 4 & 27.5 & 11 & 27.5 & 5 & 29.0 & 6 & 30.2 & 12 & 26.8 & 8\\
         & Gemini 1.5 Pro~\citep{gemini1.5} & - & 30.8 & 3 & \underline{31.4} & 2 & 31.3 & 3 & \underline{37.5} & 2 & 27.7 & 4 & 20.5 & 18 & 33.0 & 8 & 30.8 & 3 \\
         & Reka Core~\citep{team2024reka} & 67B & 26.7 & 5 & 27.7 & 4 & 26.4 & 13 & 22.5 & 14 & 26.5 & 11 & 24.0 & 15 & 34.3 & 6 & 26.9 & 7 \\
         & Reka Flash~\citep{team2024reka} & 21B & 25.5 & 8 & 24.1 & 11 & 27.2 & 11 & 30.0 & 6 & 27.5 & 5 & \underline{31.5} & 2 & 24.1 & 16 & 26.3 & 13\\
         & Reka Edge~\citep{team2024reka} & 7B & 23.8 & 15 & 20.5 & 18 & 26.3 & 16 & 22.5 & 14 & 25.5 & 14 & 22.5 & 17 & 36.8 & 3 & 25.0 & 18\\
         & GPT-4o visual caption~\citep{gpt4-o} & - & \underline{37.4} & 2 & 28.6 & 3 & \underline{32.3} & 2 & 27.5 & 11 & 25.5 & 14 & 23.0 & 16 & 28.9 & 13 & \underline{32.3} & 2\\
         & GPT-4o audio caption~\citep{gpt4-o} & - & \bf{38.6} & 1 & \bf{31.8} & 1 & \bf{33.6} & 1 & 32.5 & 4 & 27.5 & 5 & 25.0 & 12 & 26.1 & 14 & \bf{34.5} & 1 \\
         \bottomrule
    \end{tabular}  }
   \label{overall}
\end{table*}

\textbf{Quality Control}.
Duplicated text information that describes visual inputs will induce MLLMs to bypass the visual input to directly derive the answer by memorizing the answer from the internet-scale training dataset~\cite{chen2024we}. Inspired by this, we first ensure that our text questions' context is as simple as possible. Then we filter out those questions that have redundant images or audio clips by leveraging VLMs and audio LLMs. Specifically, we test all the curated questions with VLM: InternVL2~\citep{internvl}, Qwen2-VL~\citep{wang2024qwen2vl}, MiniCPM-V 2.5~\citep{yao2024minicpmv}, BLIP3~\citep{blip3}, and VILA1.5~\citep{lin2024vila} and audio LLM Qwen-Audio~\citep{chu2023qwenaudio}, Qwen2-Audio~\citep{chu2024qwen2audio}, SALMONN~\citep{tang2023salmonn}, and  Typhoon-Audio~\citep{manakul2024typhoonaudio}, and filter out those questions that can be solved by either of these models. In experiment, 2.54\% questions are filtered out because they are solved by all audio LLMs or VLMs

\section{Experiment}

We test various closed-source and open-source MLLMs
that accommodate the inputs of text, image/video, and audio. Experiments are conducted in the zero-shot setting to evaluate the performance of MLLMs without finetuning and few-shot prompting. The text prompts are designed as concise as possible to remove redundant information.

\subsection{Models}
We evaluate 18 models in total, 8 closed-source models, including Gemini 1.5 Flash, Gemini 1.5 Flash-8B, Gemini 1.5 Pro~\cite{gemini1.5}, Reka Core, Reka Flash, Reka Edge~\cite{team2024reka}, GPT-4o~\cite{gpt4-o} and 10 open-source models including Unifed-IO-2 L~\citep{unifedio}, Unified-IO-2 XL, Unified-IO-2 XXL, OneLLM~\cite{onellm}, PandaGPT~\cite{su2023pandagpt}, Video-llama~\cite{videollama}, VideoLLaMA2~\cite{videollama2}, AnyGPT~\cite{anygpt}, NExT-GPT~\cite{wu2023nextgpt}, VITA~\cite{fu2024vita}. We test those open-source models based on their source code and the latest checkpoint and test the closed-source models with available APIs.

Since we currently cannot access the GPT-4o API that supports simultaneous image, video, and audio inputs, we have adopted an alternative approach to evaluating GPT-4o models. The GPT-4o series includes two types of APIs: GPT-4o, which processes image and text inputs, and GPT-4o-audio, which processes audio and text inputs.
Based on these two APIs, we develop two methods to evaluate GPT-4o: (1) We use GPT-4o-audio to generate captions for audio clips, then feed the text, image/video, and audio captions into GPT-4o. We refer to this approach as the GPT-4o audio caption method.
(2) Similarly, we use GPT-4o to generate captions for images or videos, then input the text, audio, and visual captions into GPT-4o-audio. We refer to this approach as the GPT-4o visual caption method.

We set a random baseline which is 25\% for AV-Odyssey Bench with four-choice questions. When task performance is below the random baseline, it indicates that the model is unable to handle the task effectively. Consequently, if two models both perform worse than this random baseline, the performance gap between them becomes meaningless.

\subsection{Main Result Analysis}

In this section, we analyze the performance of MLLMs in our AV-Odyssey benchmark, as presented in Table \ref{overall}. Due to the space limit, detailed results and data distribution are provided in the Appendix. Our key findings are as follows:
 
\textbf{Challenging Nature of AV-Odyssey.} 
As presented in Table \ref{overall}, the average performance of most existing MLLMs is only marginally higher than 25\%—comparable to the expected accuracy of random guessing on four-choice questions. Notably, even the top-performing model in our AV-Odyssey, GPT-4o audio caption, only achieves 34.5\% accuracy. 
This result underscores the high level of challenge posed by our benchmark, which significantly goes beyond the distribution of training data of current models. By setting rigorous standards, our benchmark serves as a crucial tool for evaluating the robustness and versatility of MLLMs in audio-visual tasks. It highlights the limitations of existing models and provides directions for future improvements.

\textbf{Discrepancies Between Closed-Source and Open-Source Models.} The gap between open-source and closed-source models is relatively small. For example, the leading open-source models PandaGPT and OneLLM, with accuracies of 27.4\% and 27.2\% respectively, are only marginally behind GPT-4o audio caption and Gemini 1.5, which achieve accuracies of 34.5\% and 30.8\%. Besides, it can be found that PandaGPT and OneLLM deliver performance comparable to certain closed-source models, such as Gemini 1.5 Flash 8B and Reka Core. This indicates that our AV-Odyssey is challenging for both open-source and closed-source models.


\textbf{Comparison Between Audio Captions and Visual Captions.} 
It can be observed that GPT-4o audio caption achieves higher performance than GPT-4o visual caption as shown in Table \ref{overall}. This demonstrates that audio captions enable GPT-4o to process audio-visual information more effectively than visual captions.
This advantage may be due to the greater information loss in visual captions compared to audio captions when handling audio-visual content.

\label{sec:finding3}
\textbf{Limitations of Open-Source MLLMs in Connecting Audio and Visual Information.} The Audiocaps dataset~\citep{kim2019audiocaps} introduces audio captions, i.e., audio-text pairs (including animal sound audios), which are used to train MLLMs for learning audio recognition abilities by OneLLM~\citep{onellm}, Unified-IO-2~\citep{unifedio}, VideoLLaMA2~\citep{videollama2}, and NExT-GPT~\citep{wu2023nextgpt}. In addition, they adopt the image-text paired data for training vision comprehension ability. The results on our AV-Odyssey Bench showcase that the audio-text-vision training pipeline is insufficient to bridge audio and vision modalities and truly learn the audio-visual information integration capability.


\begin{figure}
\centering
\includegraphics[width=\linewidth]{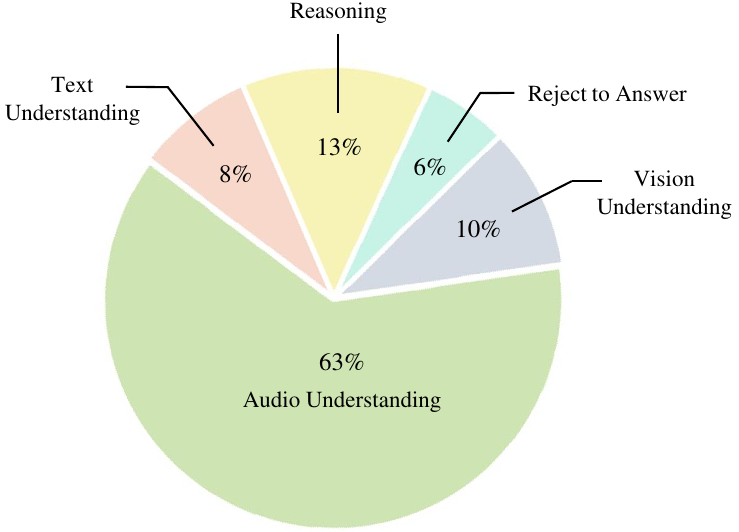}
\caption{Distribution of 104 human-annotated errors in the Gemini 1.5 Pro.}
\vspace{-0.1in}
\label{fig:error_distribution}
\vspace{-0.1in}
\end{figure}

\begin{figure}
\centering
\includegraphics[width=0.85\linewidth]{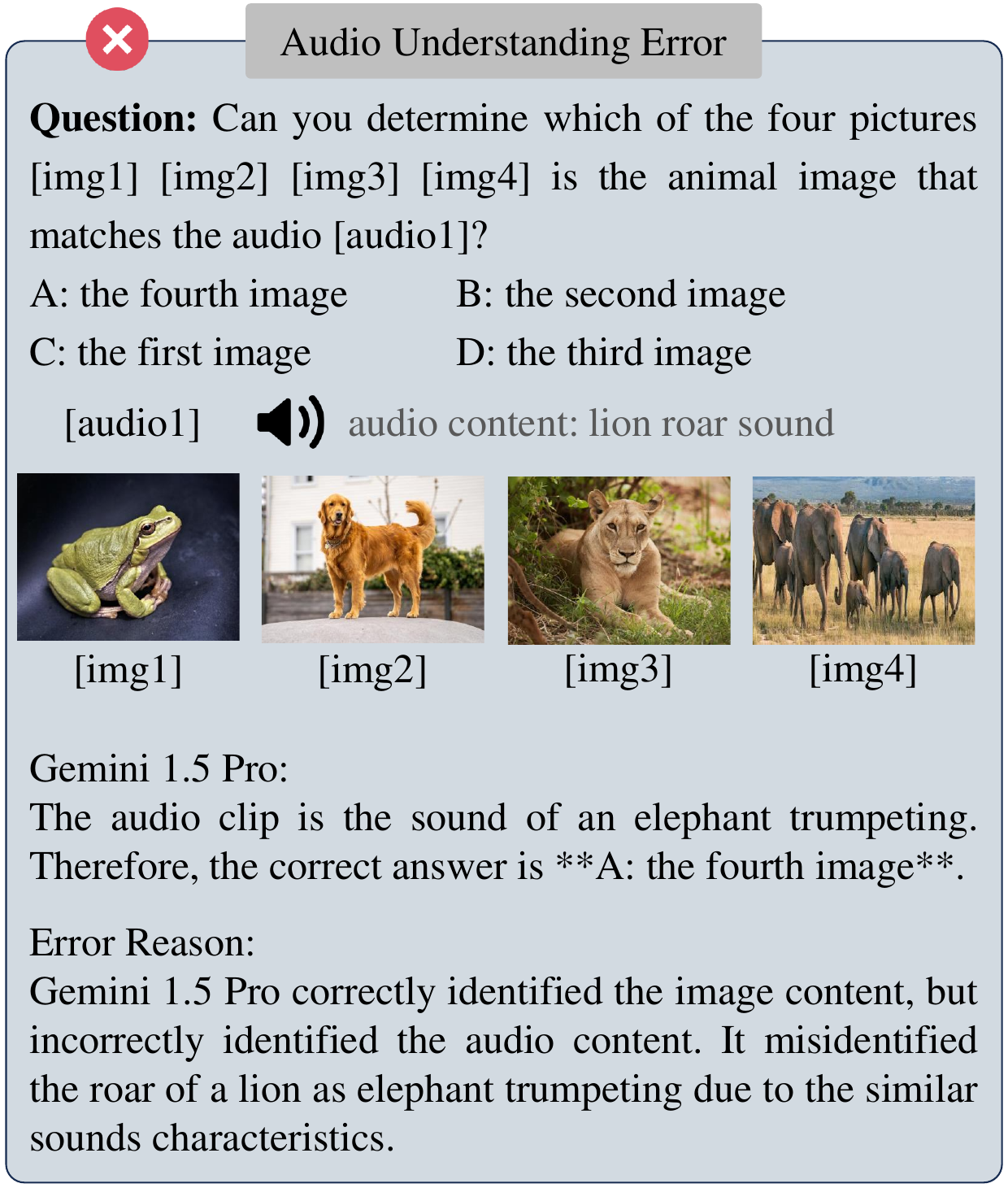}
\vspace{-0.1in}
\caption{An example of audio understanding error. More examples are provided in the Appendix.}
\label{fig:error_demo}
\vspace{-0.2in}
\end{figure}

\subsection{Error Analysis}
\label{sec:error_analysis}
In this section, we focus on the errors of Gemini 1.5 Pro to analyze the underlying causes. 
For each task, we randomly select 4 error instances for human experts to annotate the error reasons, resulting in a total of 104 instances with human-annotated error reasons.
The distribution of these errors is illustrated in Figure~\ref{fig:error_distribution}. Detailed analyses for each case are provided in the Appendix.

\textbf{Perception Understanding Errors (81\%).} \
Perception understanding errors, which include \textbf{audio understanding errors (63\%)}, vision understanding errors (10\%), and text understanding errors (8\%), form the majority of the errors.  Audio Understanding Errors stand out among all the error types with an error rate of 63\%. An example is illustrated in Figure~\ref{fig:error_demo}, where the content of the audio clip is misidentified, leading to an incorrect answer. This result indicates that the major bottleneck of audio-visual information integration is still in the perception ability in audio. This is in line with the hypothesis induced by the DeafTest results that if an MLLM has a shortage in fundamental listening ability, it will struggle in audio-visual information integration.

\textbf{Reasoning Errors (13\%).} \
In these cases, Gemini 1.5 Pro successfully extracts information from both audio and visual inputs but still produced an incorrect answer due to incorrect reasoning.

\textbf{Other Errors (6\%).} \
Other errors, primarily due to rejected answers, arise from various factors. For instance, content may be mistakenly flagged for security reasons, preventing the model from providing an answer.

\section{Conclusion}
In this work, we introduce AV-Odyssey Bench, a comprehensive audio-visual benchmark designed to evaluate the capabilities of MLLMs in understanding audio-visual information. 
Our AV-Odyssey Bench includes 4,555 meticulously crafted multiple-choice problems, each designed to challenge models in integrating information from both visual and audio cues. Through benchmarking a range of closed-source and open-source models, we uncover the current limitations of MLLMs in effectively understanding audio-visual inputs.
We hope that AV-Odyssey Bench will serve as a valuable resource for the community, facilitating the advancement of MLLMs and ultimately leading to more powerful and human-like audio-visual understanding.

{
    \small
    \bibliographystyle{ieeenat_fullname}
    \bibliography{main}
}

\clearpage
\onecolumn
\setcounter{page}{1}
\maketitlesupplementary
\appendix

\DoToC


\section{Data Distribution}

In this section, we present the detailed data distribution of our AV-Odyssey Bench in Table \ref{taskdata}. Our AV-Odyssey bench consists of 26 tasks covering a wide range of task categories.

\noindent \textbf{We will make all the data and evaluation codes public.}

\begin{table}[h]
    \centering
    \caption{Detailed task statistics in AV-Odyssey Bench.} \label{tab:timbre_task_performance}
    \small
    \resizebox{\linewidth}{!}{
\begin{tabular}{@{}ccccc@{}}
\toprule
Task ID & Task Name                    & Task Category & Class & Number \\ \midrule
1       & Instrument Recognition       & Timbre        & 28     & 200    \\
2       & Singer Recognition           & Timbre        & 20     & 200    \\
3       & Gunshot Recognition          & Timbre        & 13     & 200    \\
4       & Bird Recognition             & Timbre        & 39     & 200    \\
5       & Animal Recognition           & Timbre        & 13     & 200    \\
6       & Transportation Recognition   & Timbre        & 8      & 200    \\
7       & Material Recognition         & Timbre        & 10     & 200    \\
8       & Scene Recognition            & Timbre        & 8      & 200    \\
9       & Hazard Recognition           & Timbre        & 8      & 108    \\
10      & Action Recognition           & Timbre        & 20     & 196    \\
11      & Eating Sound Recognition     & Timbre        & 20     & 200    \\
12      & Speech Sentiment Analysis    & Tone          & 7      & 200    \\
13      & Meme Understanding           & Tone          & N/A    & 20     \\
14      & Music Sentiment Analysis     & Melody        & 7      & 197    \\
15      & Music Genre Classification   & Melody        & 8      & 200    \\
16      & Dance and Music Matching     & Melody        & 10     & 200    \\
17      & Film and Music Matching      & Melody        & 5      & 200    \\
18      & Music Score Matching         & Melody        & N/A    & 200    \\
19      & Audio 3D Angle Estimation    & Space         & N/A    & 20     \\
20      & Audio Distance Estimation    & Space         & N/A    & 20     \\
21      & Audio Time Estimation        & Time          & N/A    & 200    \\
22      & Audio-Visual Synchronization & Time          & N/A    & 200    \\
23      & Action Sequencing            & Time          & N/A    & 200    \\
24      & Hallucination Evaluation     & Hallucination & 19     & 200    \\
25      & Action Prediction            & Intricacy     & N/A    & 199    \\
26      & Action Tracing               & Intricacy     & N/A    & 195    \\ \bottomrule
\label{taskdata}
\end{tabular}
}
\end{table}
\clearpage
\onecolumn

\section{Breakdown Results}

In this section, we provide detailed results of evaluated methods on our proposed AV-Odyssey Bench, as demonstrated in Table \ref{timbretask} and Table \ref{othertask}.

\begin{table*}[ht]
    \centering
    \caption{Evaluation results of various MLLMs in `Timbre' part of AV-Odyssey Bench. The best (second best) is in bold (underline). The corresponding brackets for each task indicate the number of associated questions.} \label{tab:timbre_task_performance}
    \resizebox{\textwidth}{!}{
    \begin{tabular}{cccccccccccccc}
         \toprule
         \multicolumn{2}{c}{\multirow{2}{*}{Model}} & \multirow{2}{*}{\makecell{LLM \\ Size}} & \makecell{Instrument \\ Recognition} & \makecell{Singer \\ Recognition} & \makecell{Gunshot \\ Recognition} & \makecell{Bird \\ Recognition} & \makecell{Animal \\ Recognition} & \makecell{Transportation \\ Recognition} & \makecell{Material \\ Recognition} & \makecell{Scene \\ Recognition} & \makecell{Hazard \\ Recognition} & \makecell{Action \\ Recognition} & \makecell{Eating Sound \\Recognition}  \\
         \cmidrule{4-14}
         &&&(200) & (200) & (200) & (200) & (200) & (200) & (200) & (200) & (108) & (196) & (200)\\
         \midrule
         \multirow{10}{*}{\rotatebox{90}{Open Source}} & Unified-IO-2 L~\citep{unifedio} & 1B & 20.5 & 22.5 & 25.5 & 18.5 & 27.0 & 26.5 & 23.0 & 28.0 & 21.3 & 20.9 & 26.5\\
         & Unified-IO-2 XL~\citep{unifedio} & 3B & 20.0 & 23.5 & 24.0 & 20.5 & 27.5 & 26.0 & 27.5 & 30.0 & 19.4 & 19.9 & 26.5 \\
         & Unified-IO-2 XXL~\citep{unifedio} & 7B & 29.5 & 24.0 & 23.5 & \bf{29.0} & 23.5 & 25.5 & \underline{30.5} & 26.5 & 23.1 & 27.0 & 25.5 \\
         & OneLLM~\citep{onellm} & 7B & 26.0 & 21.5 & 27.0 & 26.0 & 22.0 & 20.0 & 29.5 & 24.5 & 26.9 & 23.0 & 29.5\\
         & PandaGPT~\citep{su2023pandagpt} & 7B & 20.0 & 21.5 & 23.0 & 17.5 & 26.0 & 26.5 & 28.0 & 27.0 & 23.1 & 21.4 & 24.5 \\
         & Video-llama~\citep{videollama} & 7B & 22.5 & 24.5 & 27.0 & \underline{26.5} & 27.0 & 23.5 & 28.0 & 25.0 & 25.0 & 26.0 & 25.5\\
         & VideoLLaMA2~\citep{videollama2} & 7B & 22.5 & 24.0 & 27.0 & 17.0 & 23.5 & 27.5 & 26.5 & 26.5 & 19.4 & 23.0 & 25.5\\
         & AnyGPT~\citep{anygpt} & 7B & 22.5 & 28.5 & 28.0 & 17.5 & 24.0 & 25.5 & 23.0 & 28.0 & 25.9 & 20.4 & 27.5 \\
         & NExT-GPT~\citep{wu2023nextgpt} & 7B & 21.0 & 23.5 & 25.5 & 21.5 & 25.5 & 25.5 & 21.0 & 24.0 & 19.4 & 23.0 & 24.0\\
         & VITA~\citep{fu2024vita} & 8 $\times$ \text{7B} & 22.0 & 20.5 & 24.5 & 21.5 & 27.5 & 25.0 & 23.5 & 28.5 & 21.3 & 19.4 & 29.5 \\
         \midrule
         \multirow{8}{*}{\rotatebox{90}{Closed Source}} & Gemini 1.5 Flash ~\citep{gemini1.5} & - & 24.5 & 24.0 & 23.5 & 17.0 & 32.5 & 26.0 & 22.5 & 29.5 & 34.3 & 48.0 & 21.5\\
         & Gemini 1.5 Flash-8B~\citep{gemini1.5} & - & 16.5 & 22.5 & 24.0 & 19.0 & 28.0 & 26.5 & 27.0 & 29.0 & 26.9 & 32.7 & 24.5 \\
         & Gemini 1.5 Pro~\citep{gemini1.5} & - & 33.0 & 26.0 & \underline{29.0} & 25.0 & 25.5 & 26.0 & 29.5 & 30.0 & 38.0 & 57.7 & 22.5\\
         & Reka Core~\citep{team2024reka} & 67B & 32.5 & 20.0 & 26.5 & 25.0 & 24.0 & 27.0 & 30.0 & 27.0 & 25.0 & 34.2 & 21.5\\
         & Reka Flash~\citep{team2024reka} & 21B & 20.0 & 22.5 & 26.5 & 26.0 & 28.5 & 26.5 & 26.5 & 29.0 & 28.7 & 22.4 & 25.0 \\
         & Reka Edge~\citep{team2024reka} & 7B & 21.5 & 24.0 & \bf{30.5} & 20.0 & 19.5 & 22.5 & 20.5 & 25.5 & 25.9 & 23.5 & 29.0\\
         & GPT-4o visual caption~\citep{gpt4-o} & - & \underline{33.0} & \underline{30.5} & 24.0 & \underline{26.5} & \underline{43.0} & \bf{42.0} & \bf{32.5} & \underline{39.0} & \bf{49.1} & \bf{67.3} & \underline{30.5}\\
         & GPT-4o audio caption~\citep{gpt4-o} & - & \bf{40.0} & \bf{38.0} & 27.5 & \underline{26.5} & \bf{45.0} & \bf{42.0} & 27.0 & \bf{41.0} & \underline{42.6} & \underline{62.2} & \bf{35.5}\\
         \bottomrule
    \end{tabular}
   }
   \label{timbretask}
\end{table*}

\begin{table*}[ht]
    \centering
    \caption{Evaluation results of various MLLMs in `Time', `Melody', `Space'. `Time', `Hallucination', and `Intricacy' parts of AV-Odyssey Bench. The best (second best) is in bold (underline). The corresponding brackets for each task indicate the number of associated questions.}\label{tab:other_task_performance}
    \vspace{3pt}
      \setlength{\tabcolsep}{0.3mm}
    \resizebox{\textwidth}{!}{
    \begin{tabular}{cccccccccccccccccc}
         \toprule
         \multicolumn{2}{c}{\multirow{4}{*}{Model}} & \multirow{4}{*}{\makecell{LLM \\ Size}} & \multicolumn{2}{c}{Tone} & \multicolumn{5}{c}{Melody} & \multicolumn{2}{c}{Space} & \multicolumn{3}{c}{Time} & \multicolumn{1}{c}{Hallucination} & \multicolumn{2}{c}{Intricacy} \\
         \cmidrule(lr){4-5}
         \cmidrule(lr){6-10}
         \cmidrule(lr){11-12}
         \cmidrule(lr){13-15}
         \cmidrule(lr){16-16}
         \cmidrule(lr){17-18}
         &&& \makecell{Speech Sentiment \\ Analysis} & \makecell{Meme \\ Understanding} & \makecell{Music Sentiment \\ Analysis} & \makecell{Music Genre \\ Classification} & \makecell{Dance and Music \\ Matching} & \makecell{Film and Music \\ Matching} & \makecell{Music Score \\ Matching} & \makecell{Audio 3D Angle \\Estimation} & \makecell{Audio Distance\\  Estimation}  & \makecell{Audio Time \\ Estimation} & \makecell{Audio-Visual \\Synchronization} & \makecell{Action \\ Sequencing} & \makecell{Hallucination \\ Evaluation} & \makecell{Action \\ Prediction} & \makecell{Action \\ Tracing}
         \\
         \cmidrule(lr){4-18}
         &&& (200) & (20) & (97) & (200) & (200) & (200) & (200) & (20) & (20) & (200) & (200) & (200) & (200) & (199) & (195)\\
         \midrule
         \multirow{10}{*}{\rotatebox{90}{Open Source}} & Unified-IO-2 L~\citep{unifedio} & 1B & 24.5 & 20.0 & \underline{27.9} & 31.0 & 27.5 & 32.5 & 24.5 & 15.0 & 15.0 & 28.0 & 25.5 & 27.0 & 30.0 & 27.1 & 33.8\\
         & Unified-IO-2 XL~\citep{unifedio} & 3B & 23.0 & 25.0 & 26.9 & 30.5 & 27.0 & 31.5 & 22.5 & 30.0 & 15.0 & 26.5 & 25.5 & 24.0 & \underline{31.5} & 35.7 & 33.8 \\
         & Unified-IO-2 XXL~\citep{unifedio} & 7B & 23.0 & 20.0 & 23.9 & 31.5 & 27.5 & 24.5 & 23.5 & \bf{50.0} & 15.0 & 28.0 & 25.0 & 27.5 & 24.5 & 33.2 & 34.4 \\
         & OneLLM~\citep{onellm} & 7B & 26.0 & 20.0 & 20.8 & 23.5 & 26.5 & 18.5 & 18.0 & \underline{45.0} & 30.0 & \bf{31.5} & \bf{29.5} & 27.0 & 25.5 & \bf{41.7} & 34.9 \\
         & PandaGPT~\citep{su2023pandagpt} & 7B & 23.5 & 20.0 & 21.6 & 28.0 & 27.0 & 32.5 & 26.0 & \underline{45.0} & \bf{45.0} & 18.5 & 26.0 & 27.0 & 28.0 & 19.6 & 28.2 \\
         & Video-llama~\citep{videollama} & 7B & 23.0 & 15.0 & 25.8 & 24.0 & 20.0 & 25.0 & \underline{28.0} & \underline{45.0} & 15.0 & 28.5 & 23.5 & 26.5 & 25.0 & 28.6 & 32.8 \\
         & VideoLLaMA2~\citep{videollama2} & 7B & 26.0 & 20.0 & 26.8 & 29.0 & 25.5 & 30.5 & 20.5 & \underline{45.0} & 15.0 & 28.5 & 26.5 & 26.5 & \bf{33.0} & 28.6 & \bf{40.5} \\
         & AnyGPT~\citep{anygpt} & 7B & 25.5 & 20.0 & 23.4 & 29.5 & 25.5 & 26.0 & 26.0 & 40.0 & 15.0 & 30.5 & \underline{28.0} & 29.0 & 29.0 & 21.1 & 30.3 \\
         & NExT-GPT~\citep{wu2023nextgpt} & 7B & 21.5 & 15.0 & 23.7 & 26.0 & \bf{28.0} & 31.0 & 28.0 & \underline{45.0} & 15.0 & \bf{31.5} & 24.0 & 31.0 & 28.5 & 20.6 & 26.7\\
         & VITA~\citep{fu2024vita} & 8 $\times$ \text{7B} & 24.5 & 45.0 & 26.8 & 26.0 & 27.5 & \underline{33.5} & 24.5 & 25.0 & 20.0 & 26.5 & 25.5 & 27.0 & 31.0 & 34.2 & \underline{39.5} \\
         \midrule
         \multirow{8}{*}{\rotatebox{90}{Closed Source}} & Gemini 1.5 Flash ~\citep{gemini1.5} & - & 23.5 & 40.0 & 21.3 & 31.0 & 27.5 & 32.5 & \underline{28.0} & 30.0 & 30.0 & 27.5 & 23.5 & 25.0 & 28.5 & 27.6 & 34.9\\
         & Gemini 1.5 Flash-8B~\citep{gemini1.5} & - & 24.5 & 25.0 & 25.9 & 33.0 & 27.5 & 32.0 & 24.5 & 40.0 & 15.0 & 31.0 & 25.5 & 26.0 & 29.0 & 25.6 & 34.9 \\
         & Gemini 1.5 Pro~\citep{gemini1.5} & - & \bf{29.5} & 50.0 & 25.4 & 42.5 & \bf{28.0} & 28.5 & \bf{29.0} & 35.0 & \underline{40.0} & 30.0 & 24.5 & 28.5 & 20.5 & 32.2 & 33.8 \\
         & Reka Core~\citep{team2024reka} & 67B & \underline{28.5} & 20.0 & 22.8 & 24.5 & 27.5 & 30.0 & 25.5 & 25.0 & 20.0 & 30.0 & 25.5 & 24.0 & 24.0 & 33.7 & 34.9\\
         & Reka Flash~\citep{team2024reka} & 21B & 24.5 & 20.0 & \bf{30.5} & 29.5 & 27.5 & 25.5 & 24.5 & \underline{45.0} & 15.0 & 30.0 & 25.5 & 27.0 & \underline{31.5} & 19.1 & 29.2 \\
         & Reka Edge~\citep{team2024reka} & 7B & 20.5 & 20.0 & 24.9 & 24.5 & 27.5 & 30.0 & 24.0 & 30.0 & 15.0 & 30.0 & 25.5 & 21.0 & 22.5 & \underline{38.2} & 35.4 \\
         & GPT-4o visual caption~\citep{gpt4-o} & - & 26.0 & \underline{55.0} & 24.4 & \underline{48.0} & 27.0 & \bf{34.5} & 23.5 & 25.0 & 30.0 & 21.5 & 22.5 & \underline{32.5} & 23.0 & 32.2 & 25.6\\
         & GPT-4o audio caption~\citep{gpt4-o} & - & 28.0 & \bf{70.0} & 24.4 & \bf{56.5} & 27.5 & 32.5 & 22.5 & 30.0 & 35.0 & 23.5 & 25.5 & \bf{33.5} & 25.0 & 30.2 & 22.0\\ \bottomrule
    \end{tabular}
    }
    \label{othertask}
\end{table*}

\clearpage
\onecolumn
\section{Case Study}
\phantomsection
\label{list:list_of_figures}
\listofappfigures

\renewcommand{\thefigure}{\arabic{figure}}

\clearpage

\begin{figure*}[!htbp]
\centering
\includegraphics[width=\textwidth]{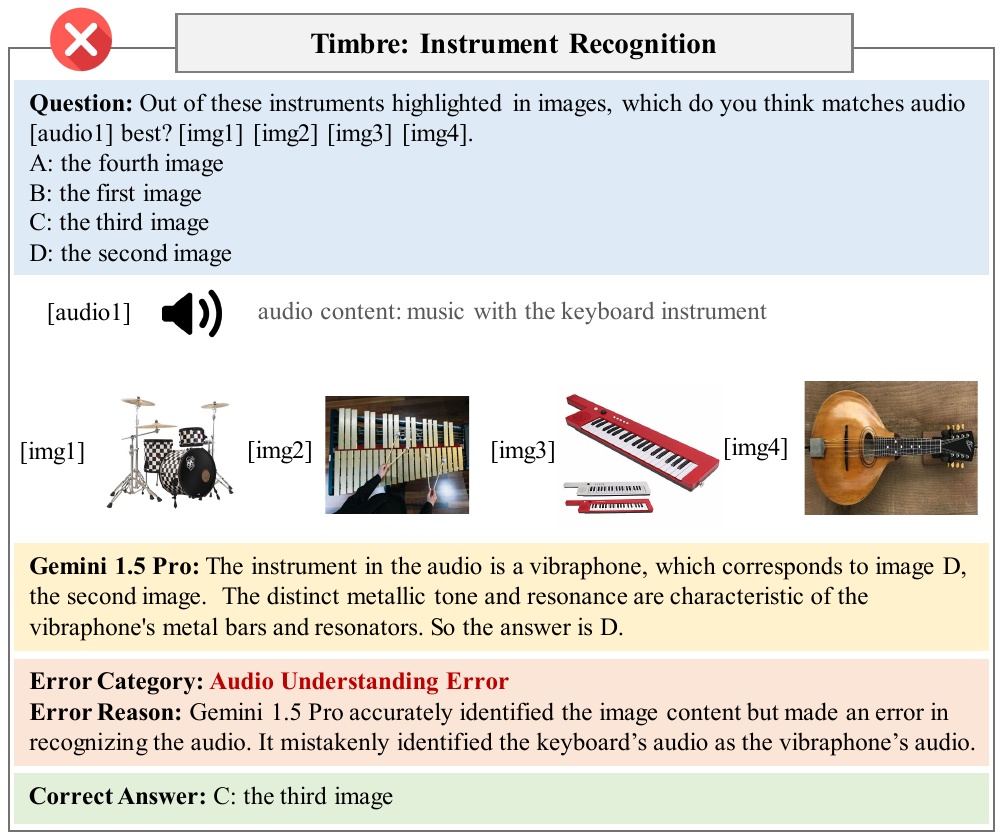}
\caption{A sampled error case in the instrument recognition task.}
\addcontentsline{afg}{appfigures}{\protect\numberline{\thefigure}Timbre, Instrument Recognition: Audio Understanding Error}
\end{figure*}
\newpage

\begin{figure*}[!htbp]
\centering
\includegraphics[width=\textwidth]{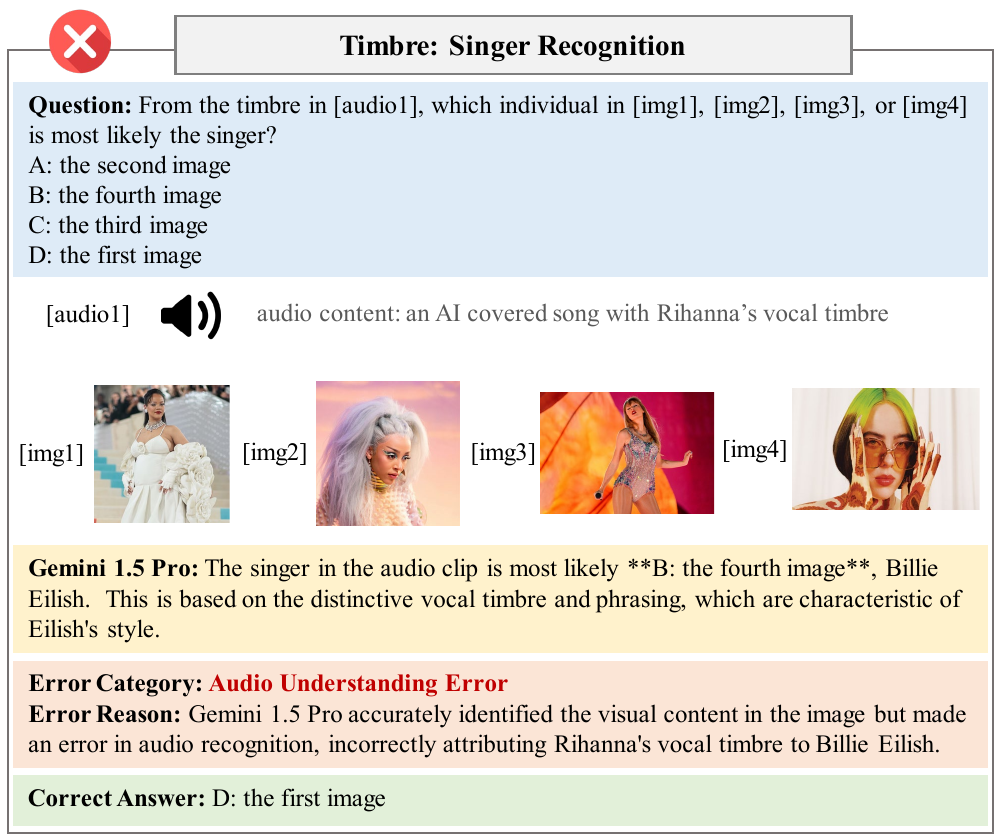}
\caption{A sampled error case in the singer recognition task.}
\addcontentsline{afg}{appfigures}{\protect\numberline{\thefigure}Timbre, Singer Recognition: Audio Understanding Error}
\end{figure*}
\newpage

\begin{figure*}[!htbp]
\centering
\includegraphics[width=\textwidth]{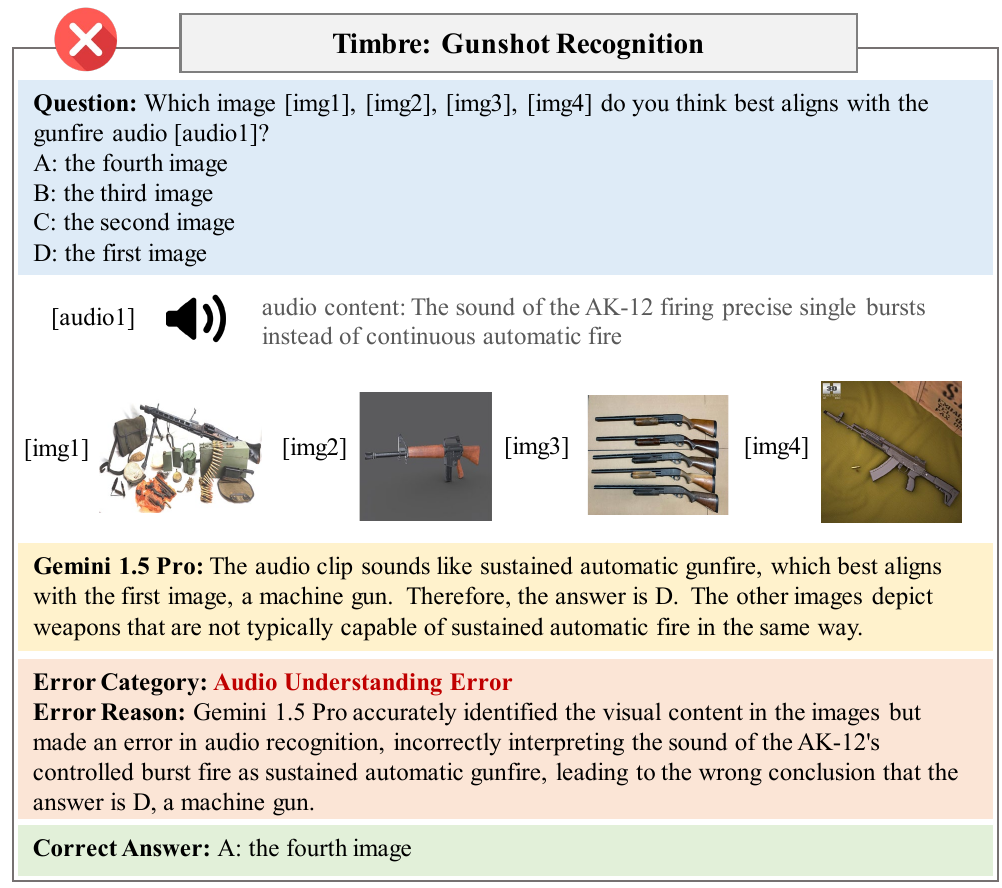}
\caption{A sampled error case in the gunshot recognition task.}
\addcontentsline{afg}{appfigures}{\protect\numberline{\thefigure}Timbre, Gunshot Recognition: Audio Understanding Error}
\end{figure*}
\newpage

\begin{figure*}[!htbp]
\centering
\includegraphics[width=\textwidth]{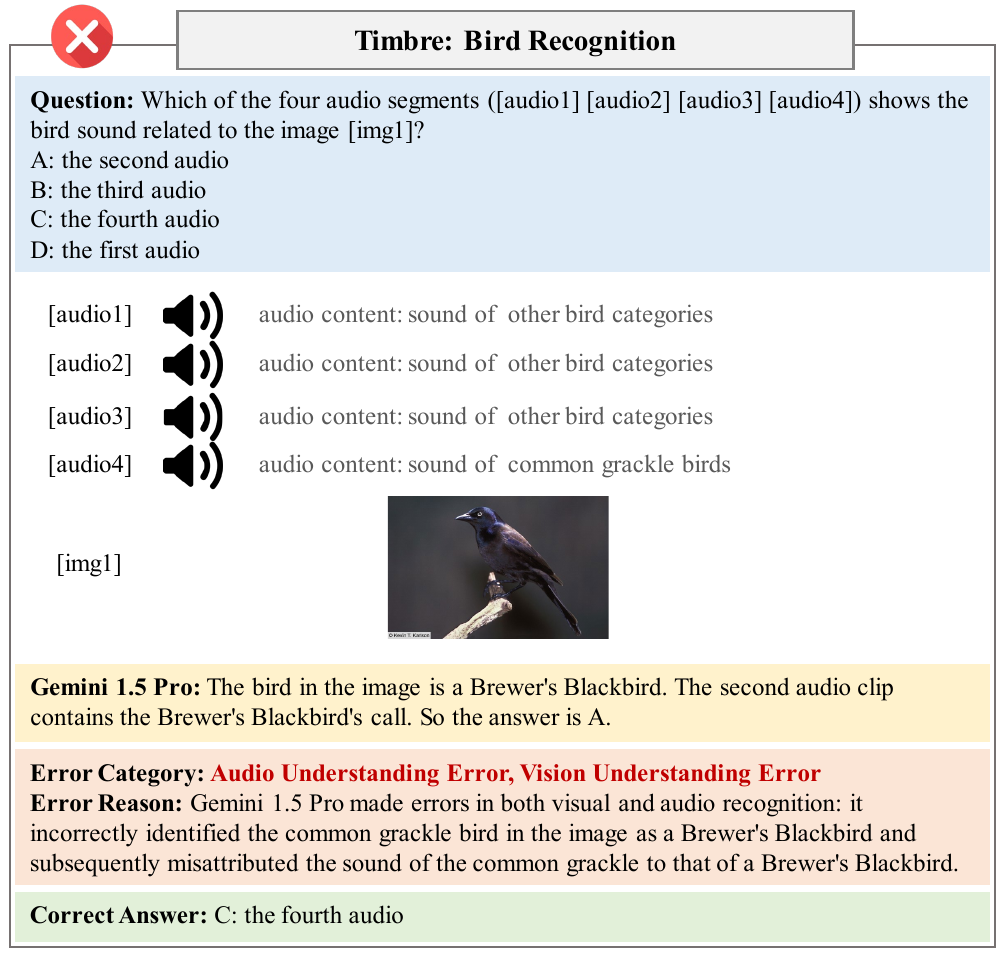}
\caption{A sampled error case in the bird recognition task.}
\addcontentsline{afg}{appfigures}{\protect\numberline{\thefigure}Timbre, Bird Recognition: Audio Understanding Error, Vision Understanding Error}
\end{figure*}
\newpage

\begin{figure*}[!htbp]
\centering
\includegraphics[width=\textwidth]{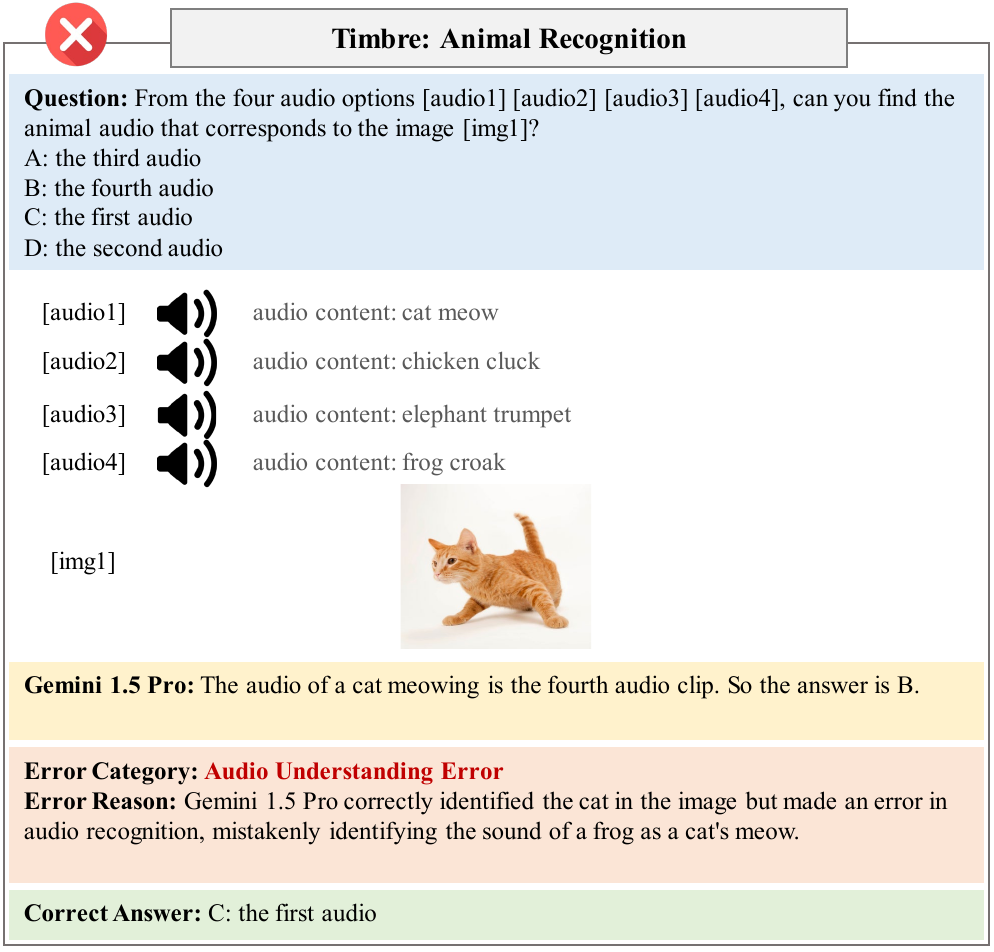}
\caption{A sampled error case in the animal recognition task.}
\addcontentsline{afg}{appfigures}{\protect\numberline{\thefigure}Timbre, Animal Recognition: Audio Understanding Error}
\end{figure*}
\newpage

\begin{figure*}[!htbp]
\centering
\includegraphics[width=\textwidth]{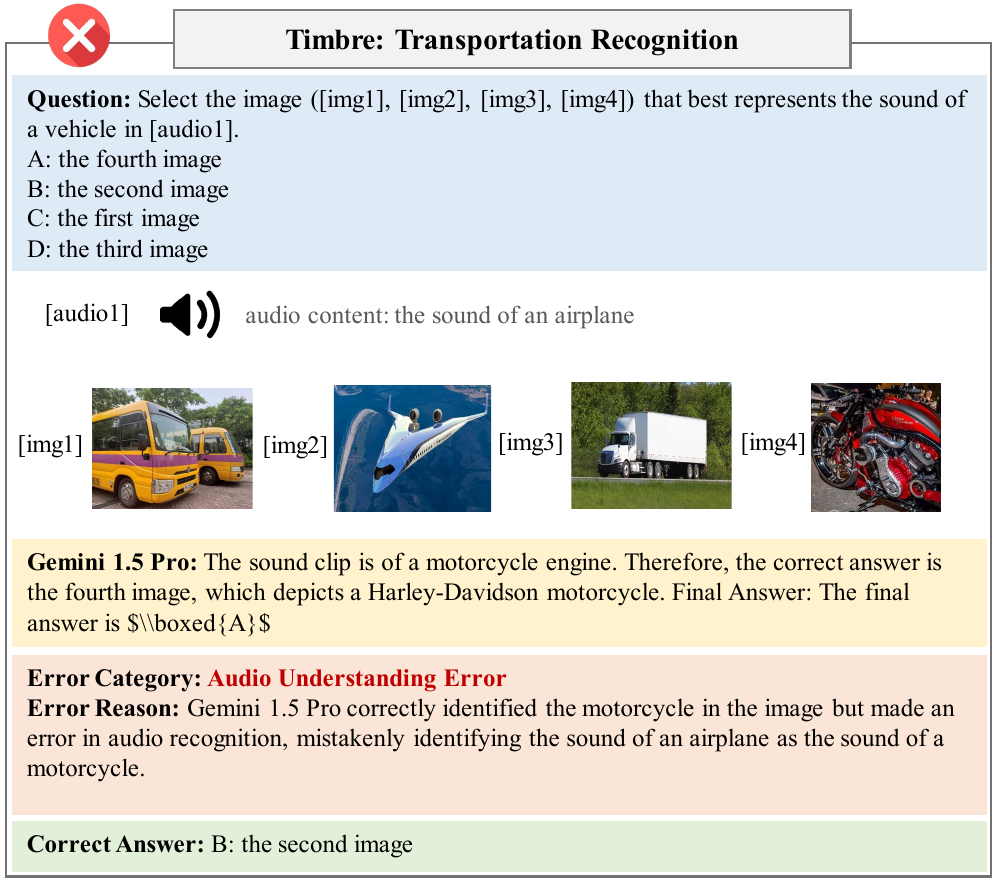}
\caption{A sampled error case in the transportation recognition task.}
\addcontentsline{afg}{appfigures}{\protect\numberline{\thefigure}Timbre, Transportation Recognition: Audio Understanding Error}
\end{figure*}
\newpage

\begin{figure*}[!htbp]
\centering
\includegraphics[width=\textwidth]{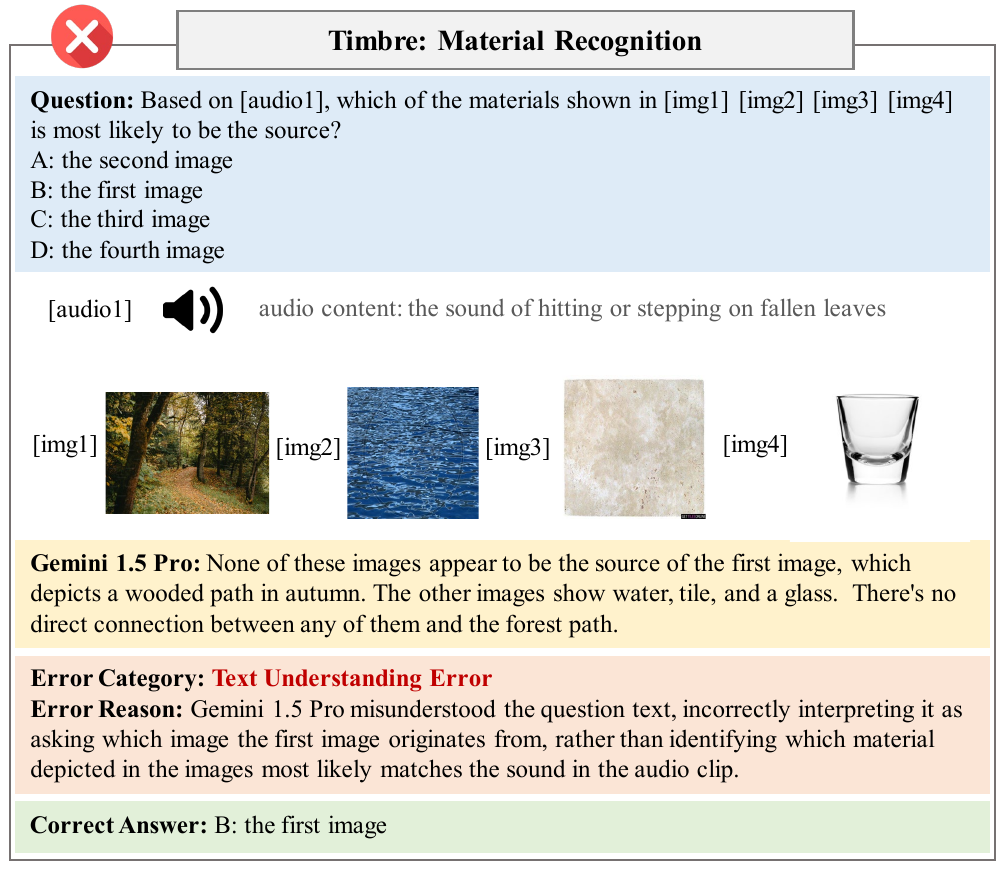}
\caption{A sampled error case in the material recognition task.}
\addcontentsline{afg}{appfigures}{\protect\numberline{\thefigure}Timbre, Material Recognition: Text Understanding Error}
\end{figure*}
\newpage

\begin{figure*}[!htbp]
\centering
\includegraphics[width=\textwidth]{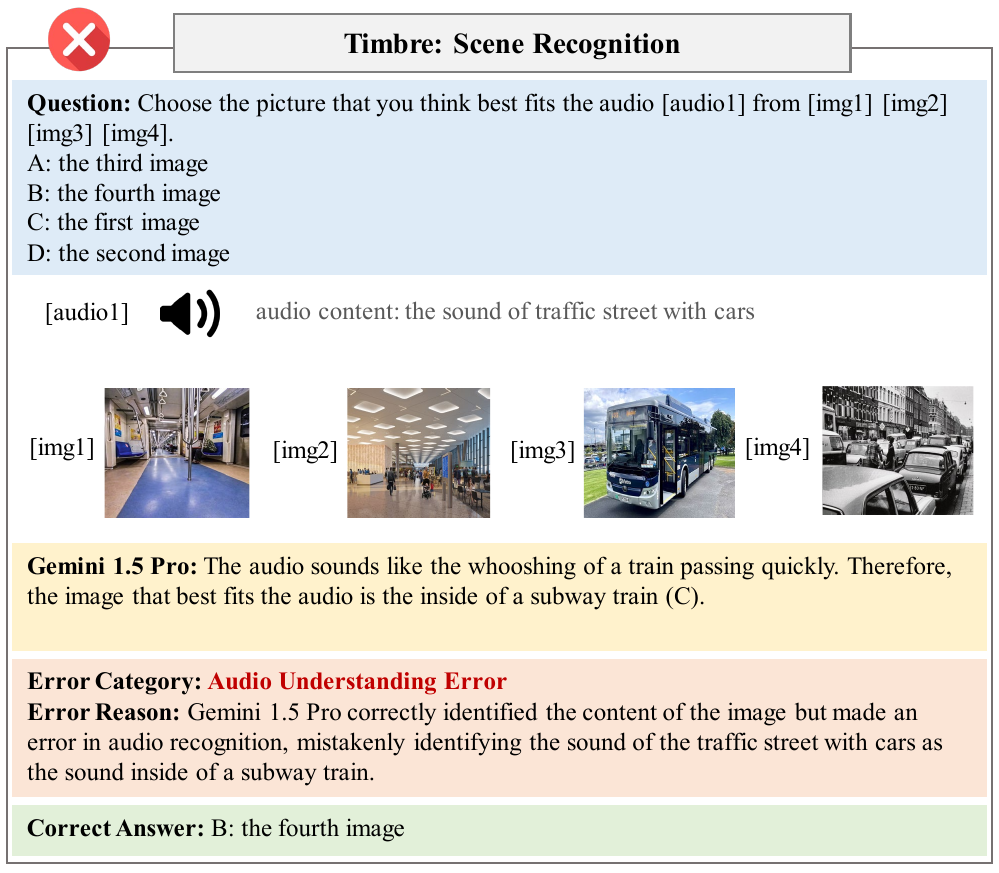}
\caption{A sampled error case in the scene recognition task.}
\addcontentsline{afg}{appfigures}{\protect\numberline{\thefigure}Timbre, Scene Recognition: Audio Understanding Error}
\end{figure*}
\newpage

\begin{figure*}[!htbp]
\centering
\includegraphics[width=\textwidth]{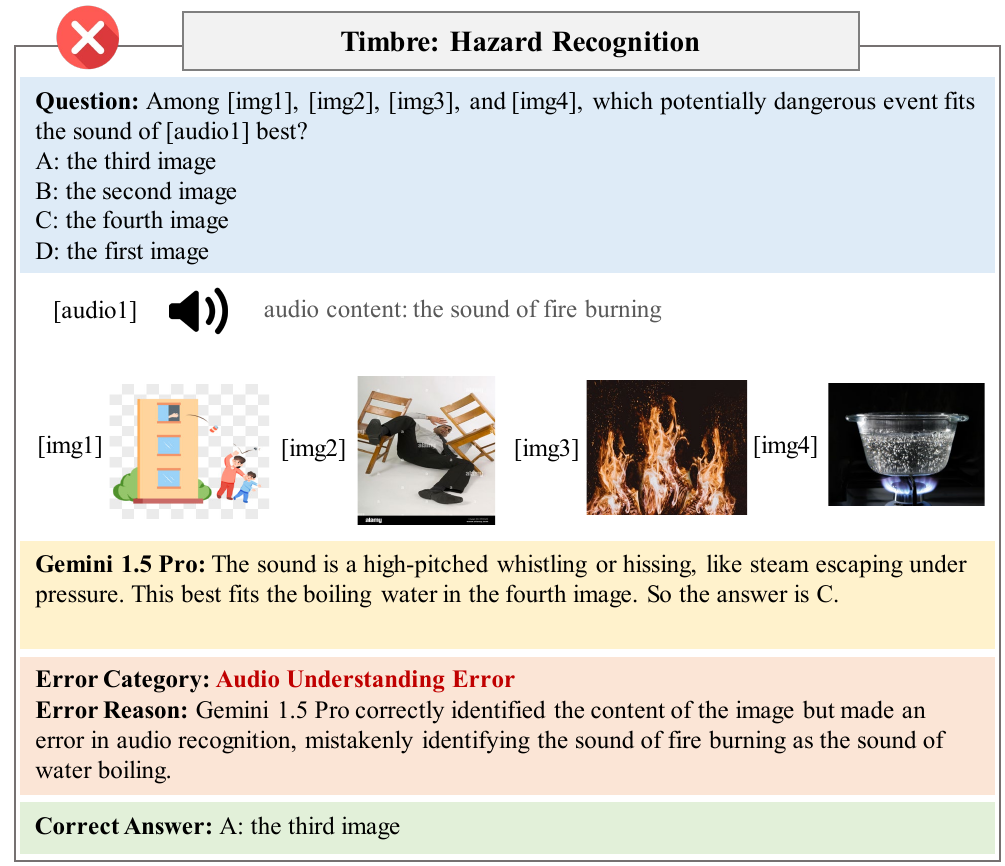}
\caption{A sampled error case in the hazard recognition task.}
\addcontentsline{afg}{appfigures}{\protect\numberline{\thefigure}Timbre, Hazard Recognition: Audio Understanding Error}
\end{figure*}
\newpage

\begin{figure*}[!htbp]
\centering
\includegraphics[width=\textwidth]{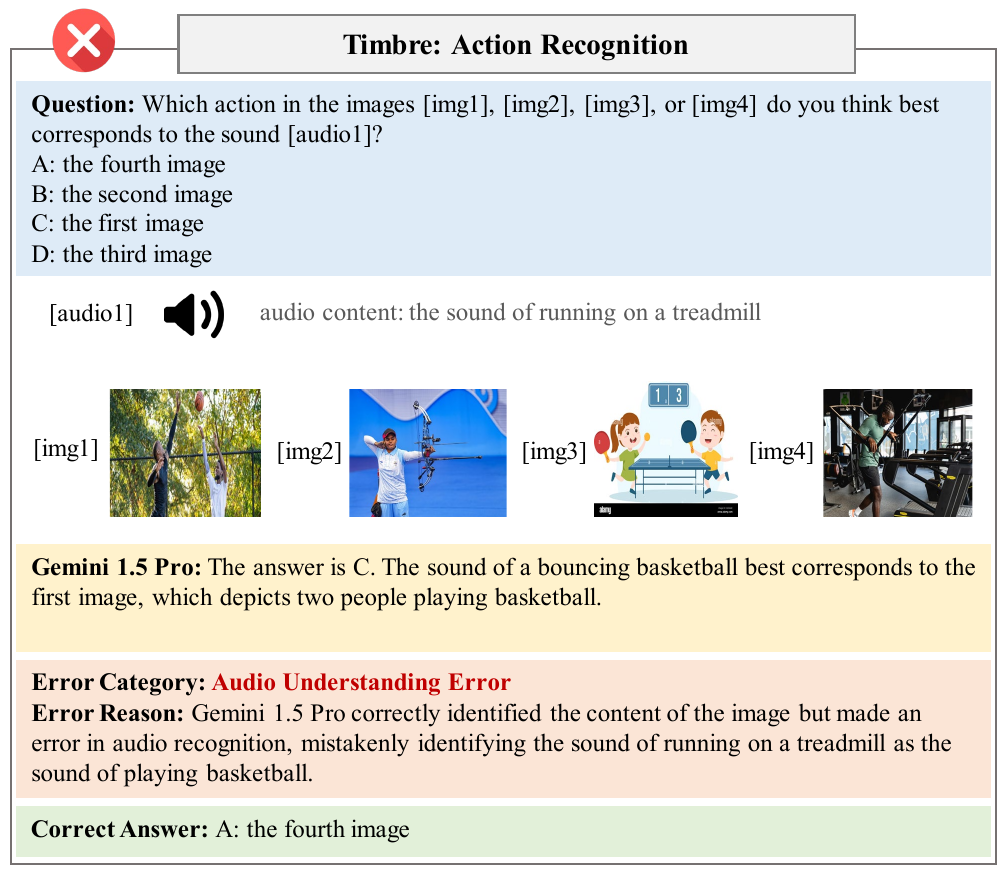}
\caption{A sampled error case in the action recognition task.}
\addcontentsline{afg}{appfigures}{\protect\numberline{\thefigure}Timbre, Action Recognition: Audio Understanding Error}
\end{figure*}
\newpage

\begin{figure*}[!htbp]
\centering
\includegraphics[width=\textwidth]{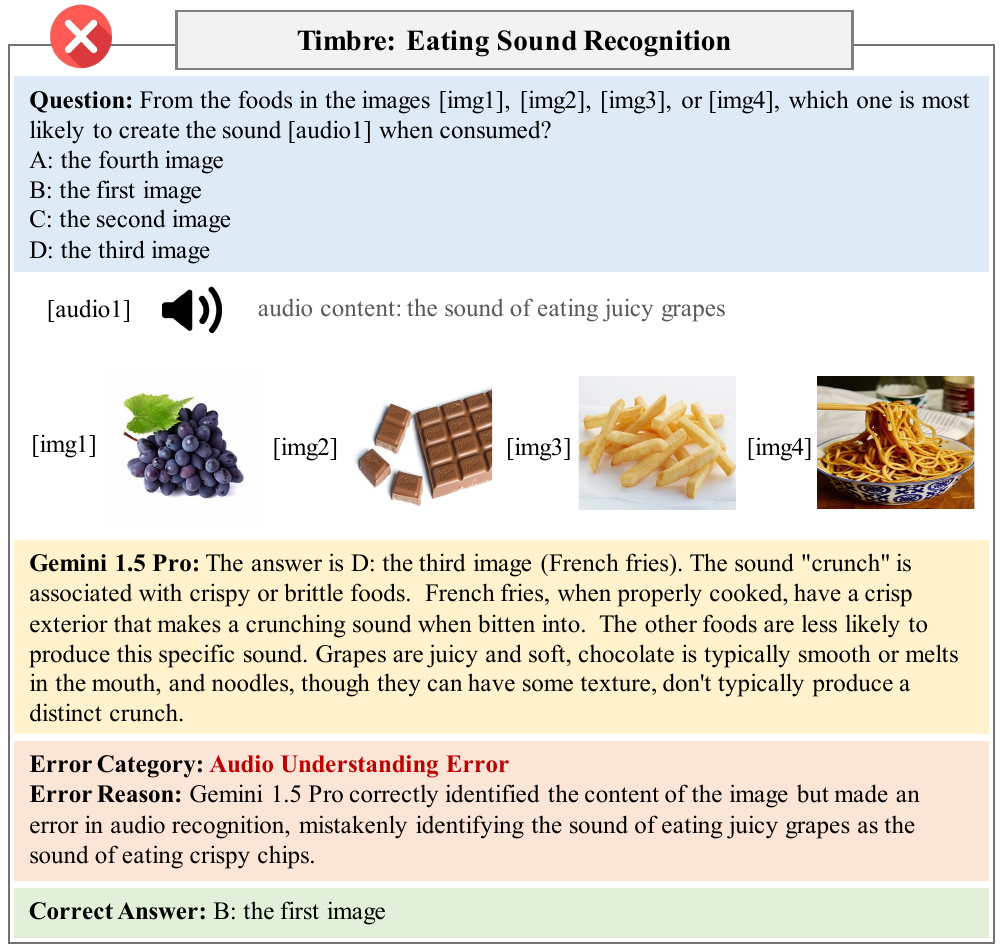}
\caption{A sampled error case in the eating sound recognition task.}
\addcontentsline{afg}{appfigures}{\protect\numberline{\thefigure}Timbre, Eating Sound Recognition: Audio Understanding Error}
\end{figure*}
\newpage

\begin{figure*}[!htbp]
\centering
\includegraphics[width=\textwidth]{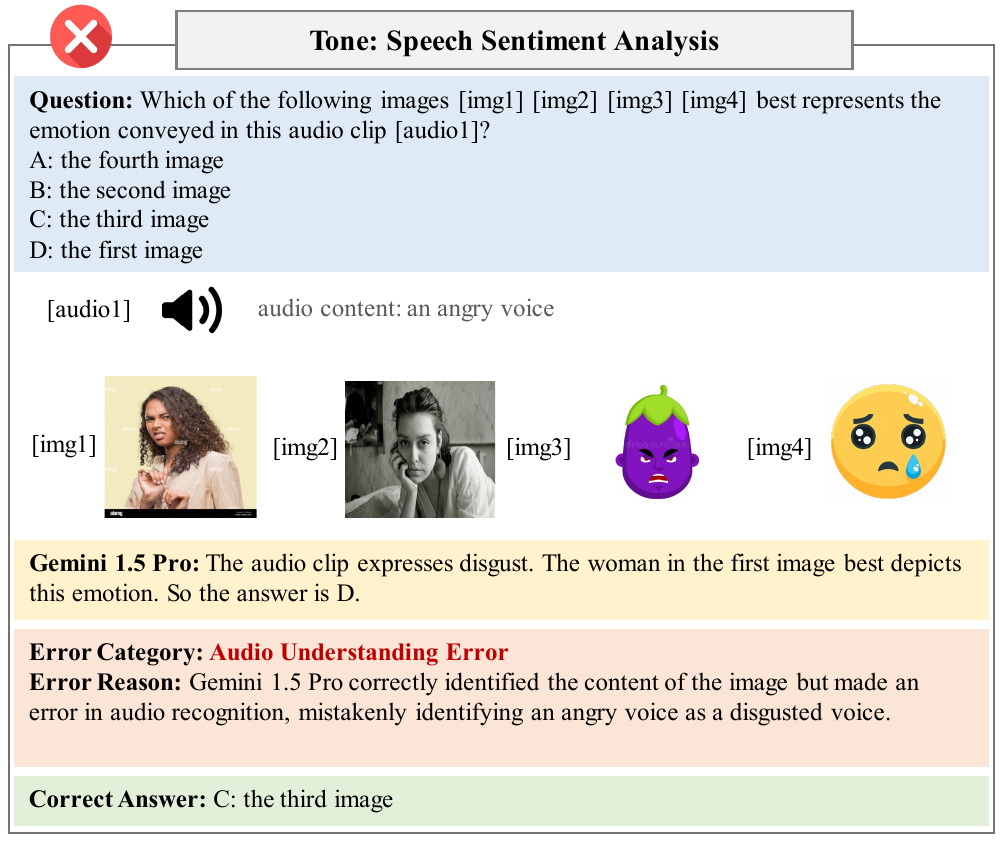}
\caption{A sampled error case in the speech sentiment analysis task.}
\addcontentsline{afg}{appfigures}{\protect\numberline{\thefigure}Tone, Speech Recognition: Audio Understanding Error}
\end{figure*}
\newpage

\begin{figure*}[!htbp]
\centering
\includegraphics[width=\textwidth]{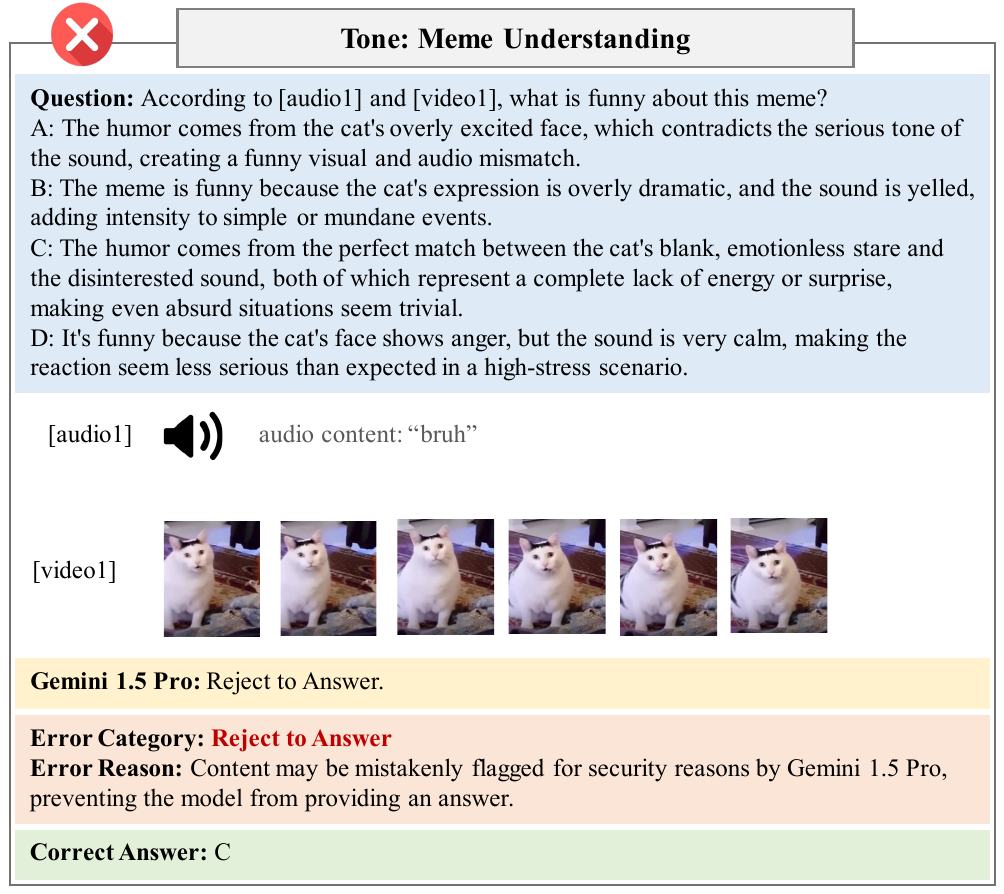}
\caption{A sampled error case in the meme understanding task.}
\addcontentsline{afg}{appfigures}{\protect\numberline{\thefigure}Tone, Meme Recognition: Reject to Answer}
\end{figure*}
\newpage

\begin{figure*}[!htbp]
\centering
\includegraphics[width=\textwidth]{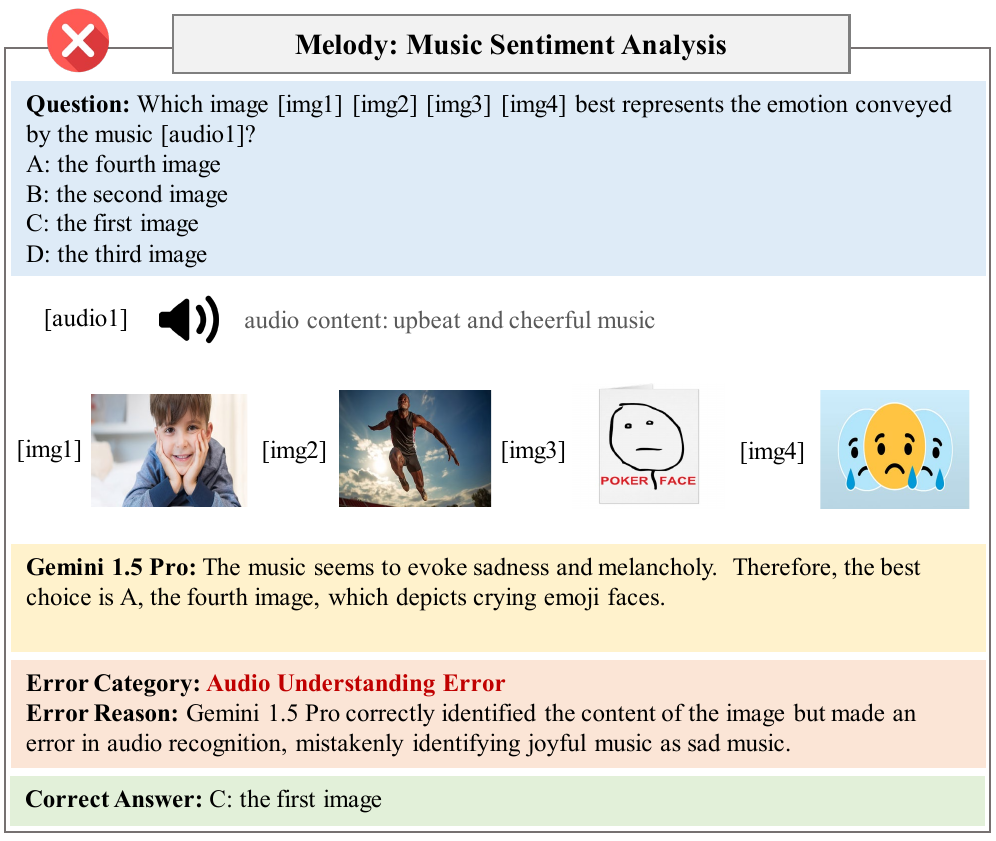}
\caption{A sampled error case in the music sentiment analysis task.}
\addcontentsline{afg}{appfigures}{\protect\numberline{\thefigure}Melody, Music Sentiment Recognition: Audio Understanding Error}
\end{figure*}
\newpage

\begin{figure*}[!htbp]
\centering
\includegraphics[width=\textwidth]{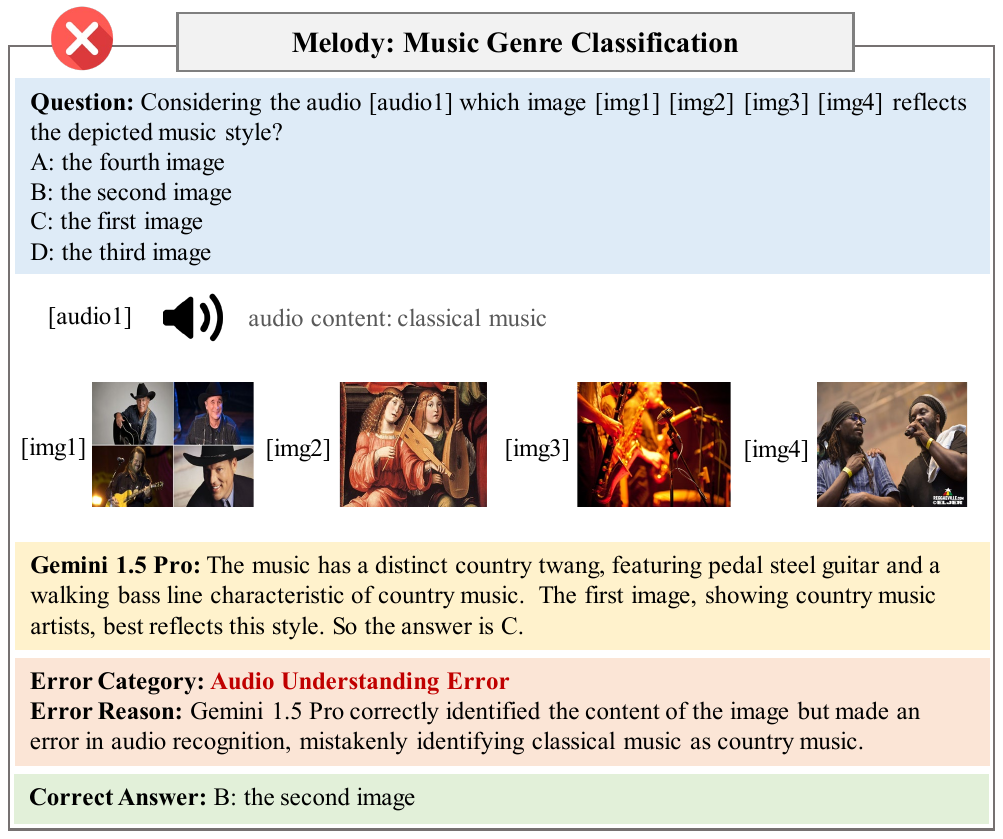}
\caption{A sampled error case in the music genre classification task.}
\addcontentsline{afg}{appfigures}{\protect\numberline{\thefigure}Melody, Music Genre Classification: Audio Understanding Error}
\end{figure*}
\newpage

\begin{figure*}[!htbp]
\centering
\includegraphics[width=\textwidth]{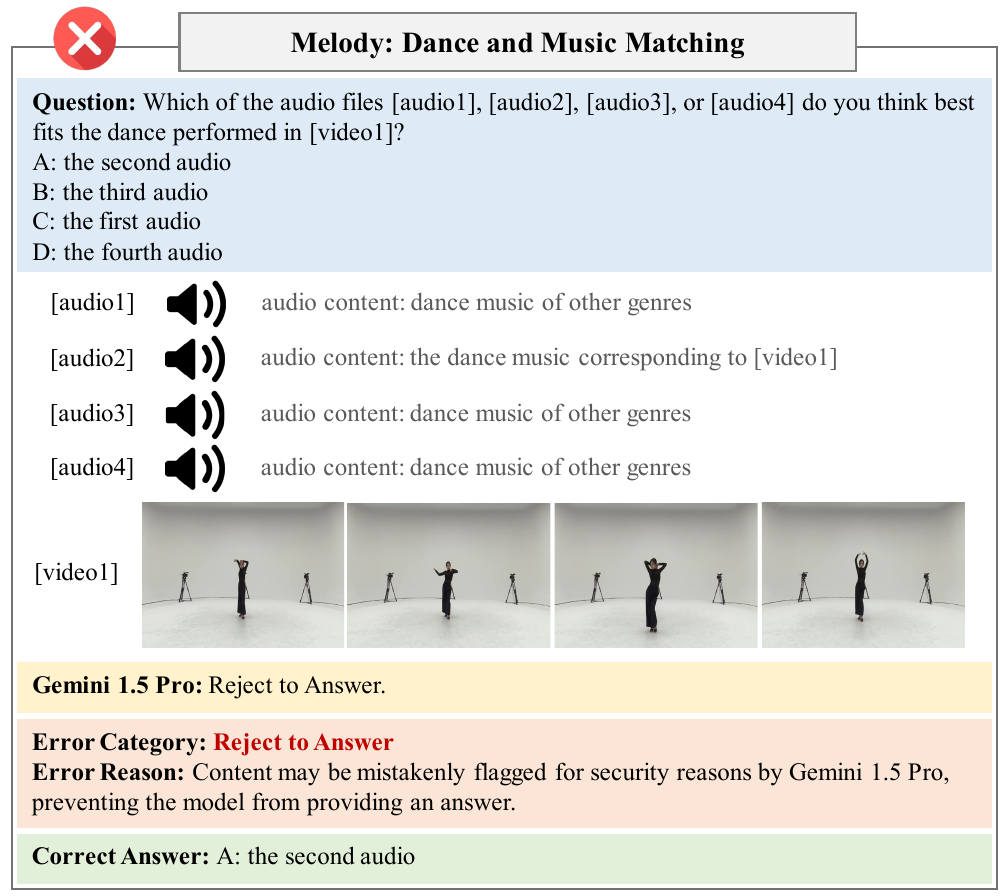}
\caption{A sampled error case in the dance and music matching task.}
\addcontentsline{afg}{appfigures}{\protect\numberline{\thefigure}Melody, Dance and Music Matching: Reject to Answer}
\end{figure*}
\newpage

\begin{figure*}[!htbp]
\centering
\includegraphics[width=\textwidth]{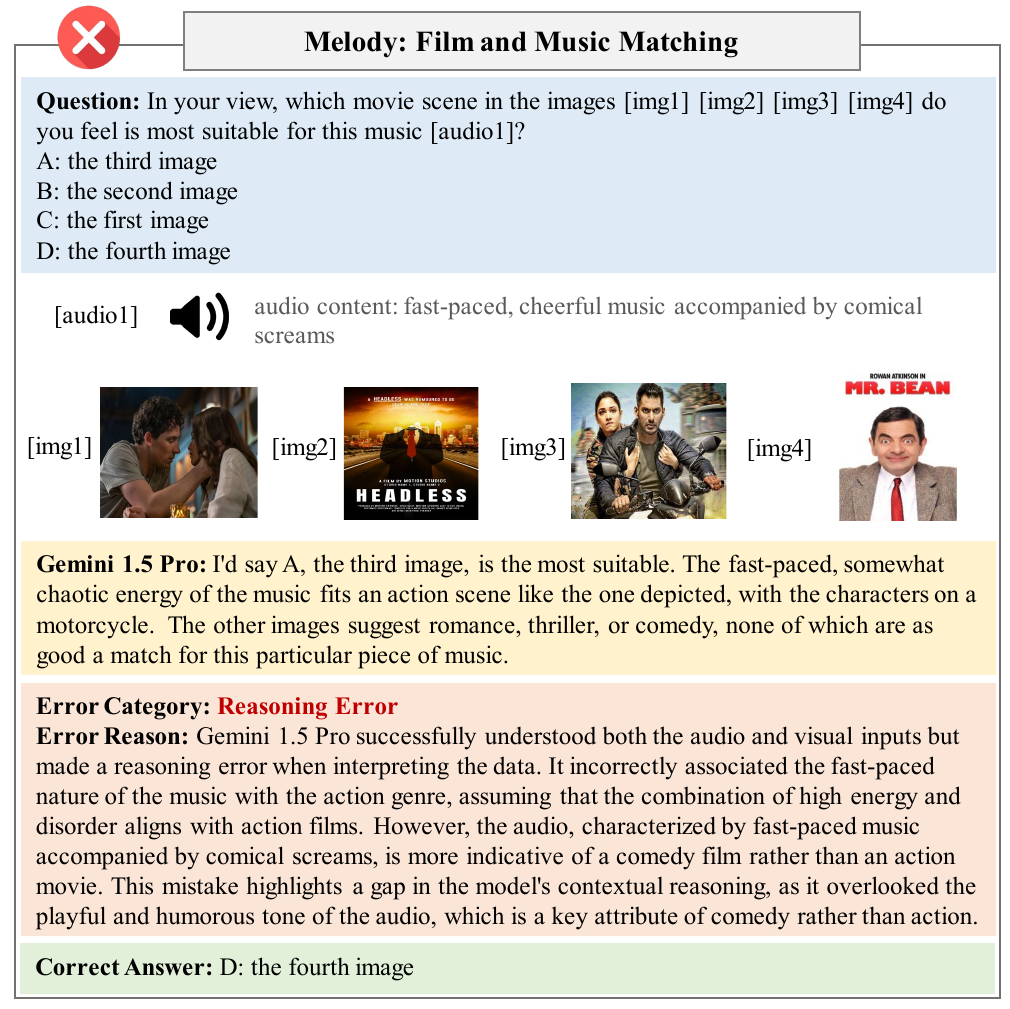}
\caption{A sampled error case in the film and music matching task.}
\addcontentsline{afg}{appfigures}{\protect\numberline{\thefigure}Melody, Film and Music Matching: Reasoning Error}
\end{figure*}
\newpage

\begin{figure*}[!htbp]
\centering
\includegraphics[width=\textwidth]{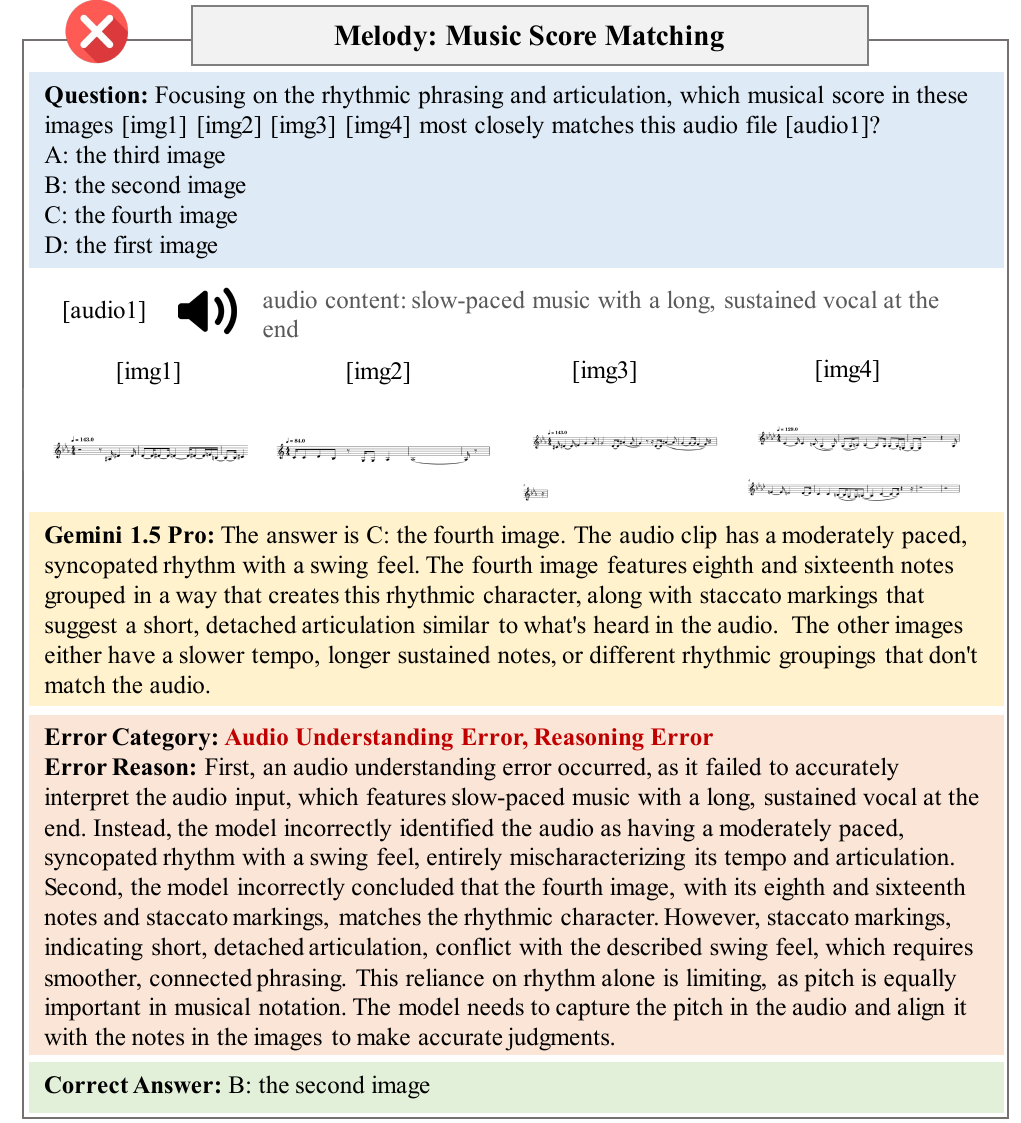}
\caption{A sampled error case in the music score matching task.}
\addcontentsline{afg}{appfigures}{\protect\numberline{\thefigure}Melody, Music Score Matching: Audio Understanding Error, Reasoning Error}
\end{figure*}
\newpage

\begin{figure*}[!htbp]
\centering
\includegraphics[width=\textwidth]{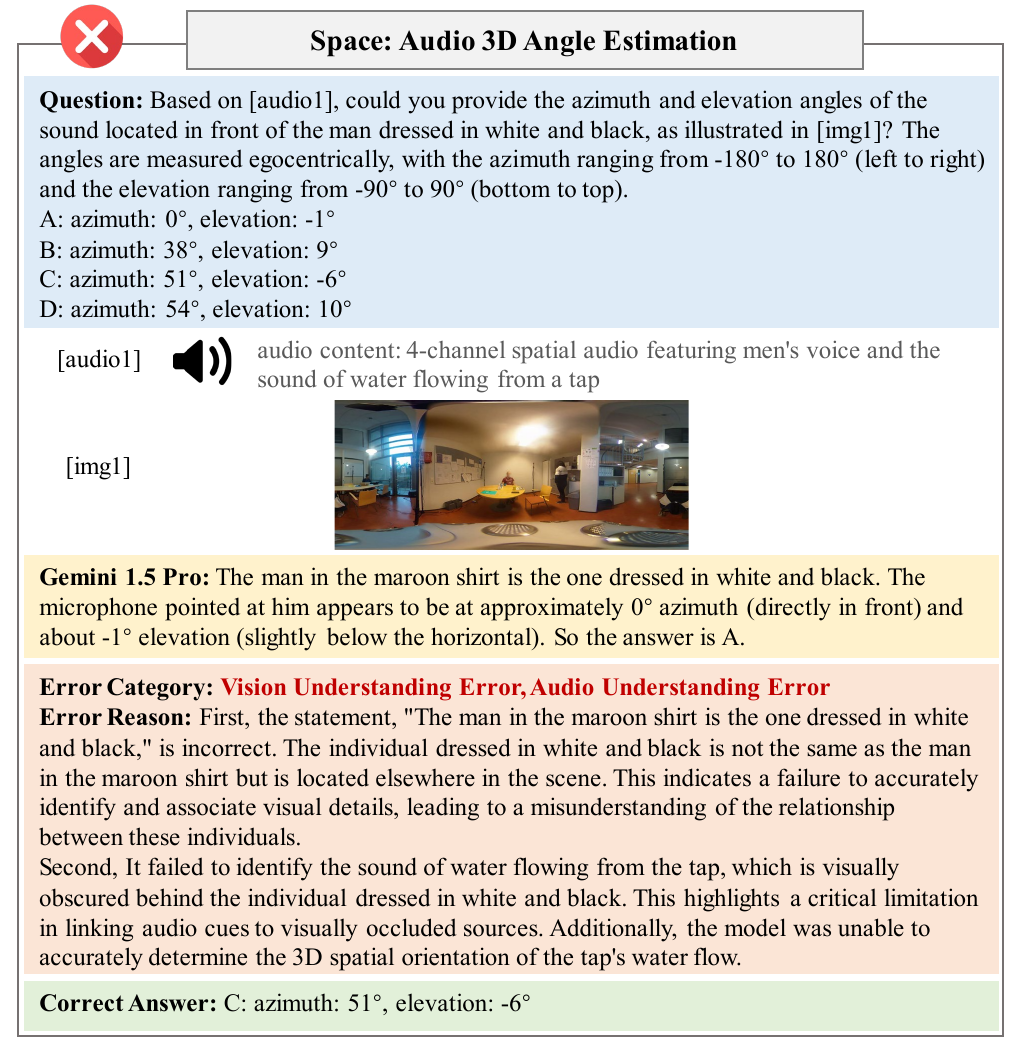}
\caption{A sampled error case in the audio 3D angle estimation task.}
\addcontentsline{afg}{appfigures}{\protect\numberline{\thefigure}Space, Audio 3D Angle Estimation: Vision Understanding Error, Audio Understanding Error}
\end{figure*}
\newpage

\begin{figure*}[!htbp]
\centering
\includegraphics[width=\textwidth]{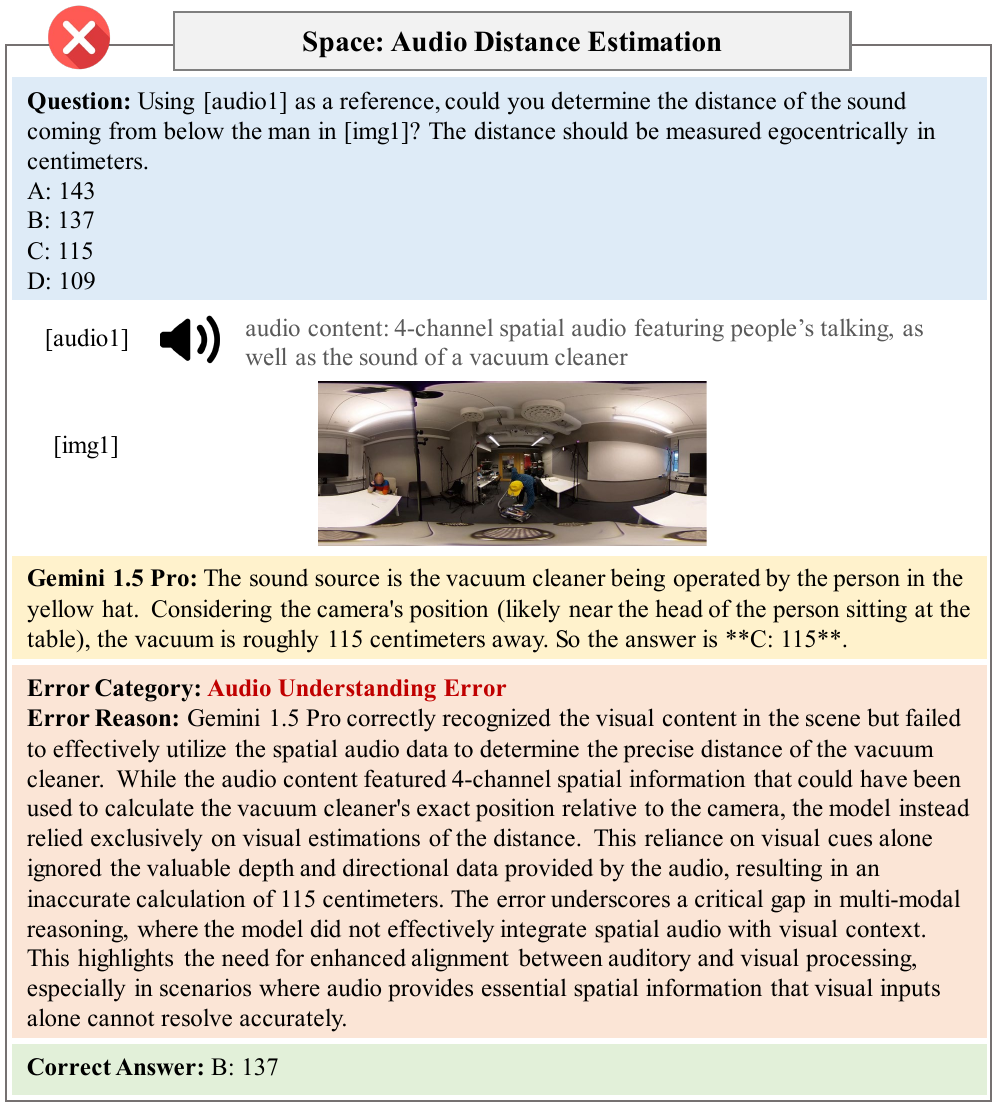}
\caption{A sampled error case in the audio distance estimation task.}
\addcontentsline{afg}{appfigures}{\protect\numberline{\thefigure}Space, Audio Distance Estimation: Audio Understanding Error}
\end{figure*}
\newpage

\begin{figure*}[!htbp]
\centering
\includegraphics[width=\textwidth]{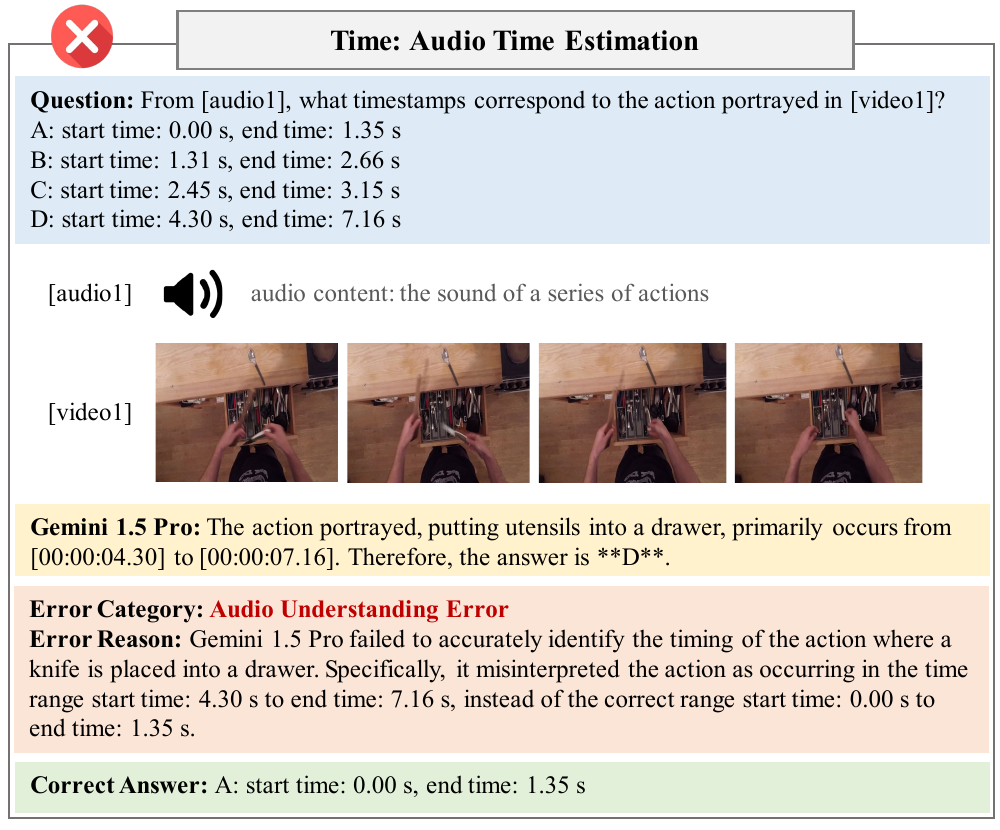}
\caption{A sampled error case in the audio time estimation task.}
\addcontentsline{afg}{appfigures}{\protect\numberline{\thefigure}Time, Audio Time Estimation: Audio Understanding Error}
\end{figure*}
\newpage

\begin{figure*}[!htbp]
\centering
\includegraphics[width=\textwidth]{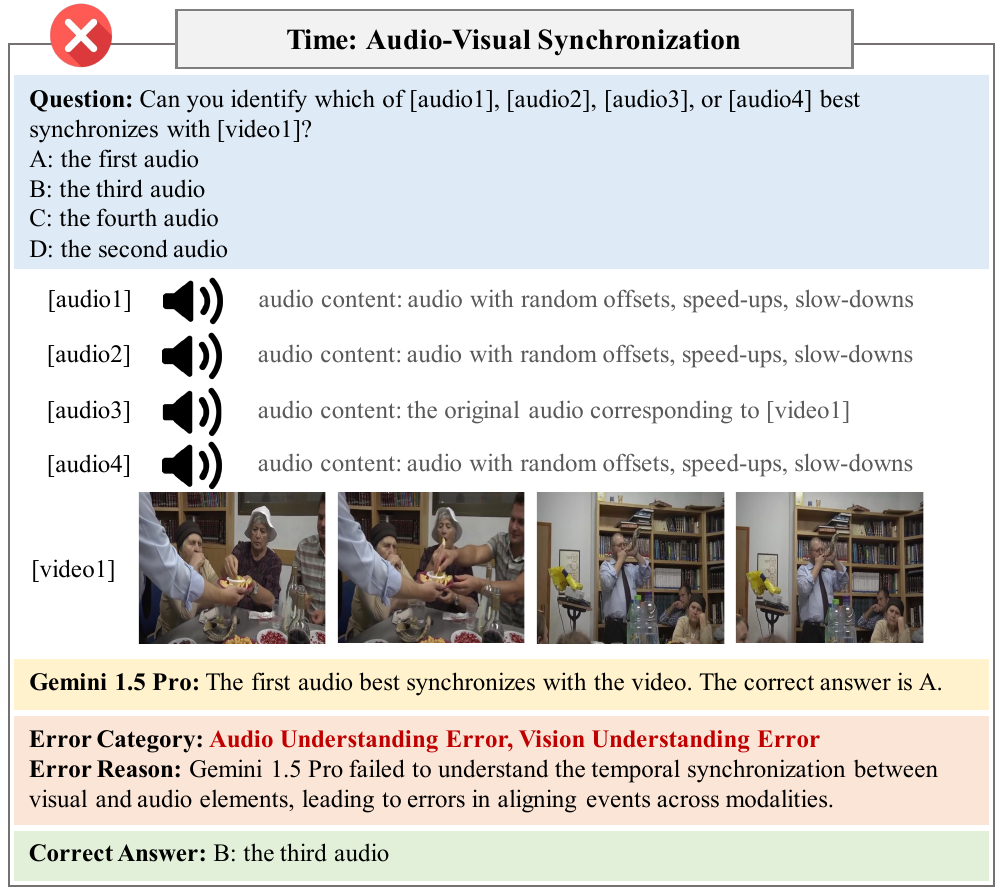}
\caption{A sampled error case in the audio-visual synchronization task.}
\addcontentsline{afg}{appfigures}{\protect\numberline{\thefigure}Time, Audio-Visual Synchronization: Audio Understanding Error, Vision Understanding Error}
\end{figure*}
\newpage

\begin{figure*}[!htbp]
\centering
\includegraphics[width=\textwidth]{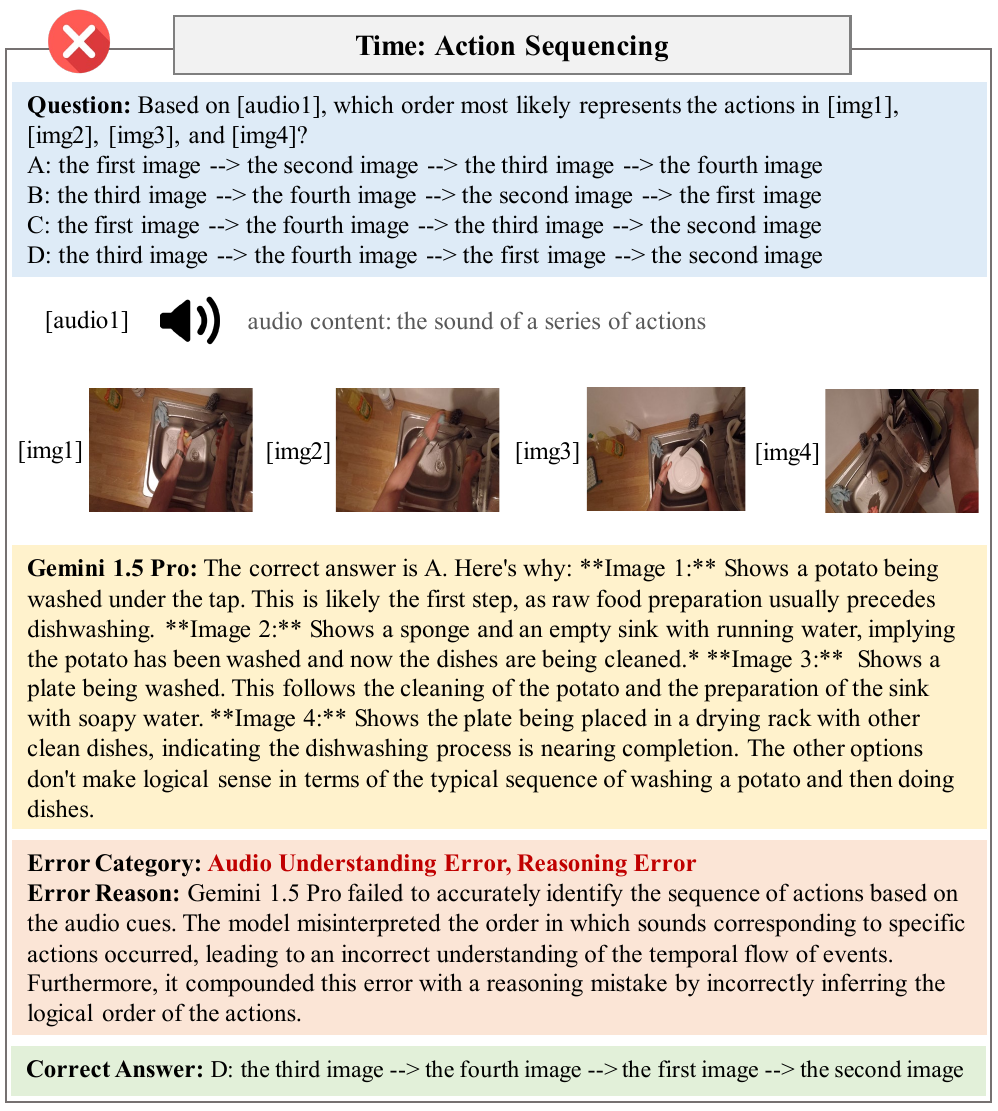}
\caption{A sampled error case in the action sequencing task.}
\addcontentsline{afg}{appfigures}{\protect\numberline{\thefigure}Time, Action Sequencing: Audio Understanding Error, Reasoning Error}
\end{figure*}
\newpage

\begin{figure*}[!htbp]
\centering
\includegraphics[width=\textwidth]{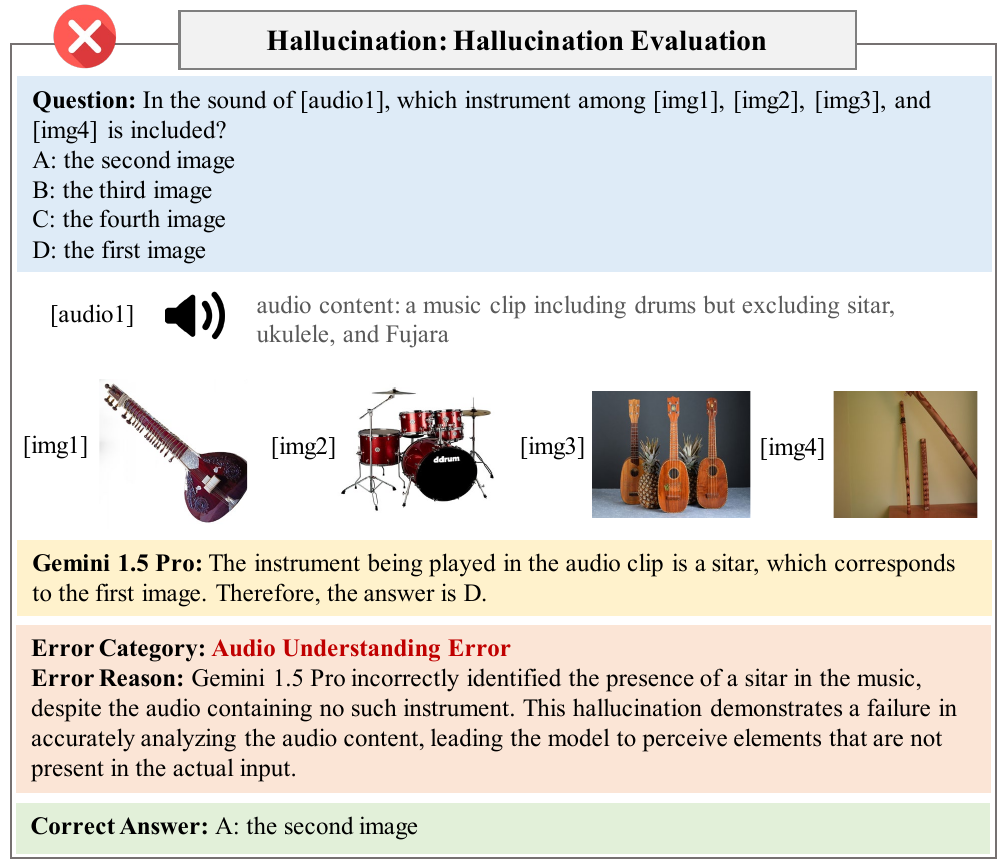}
\caption{A sampled error case in the hallucination evaluation task.}
\addcontentsline{afg}{appfigures}{\protect\numberline{\thefigure}Hallucination, Hallucination Evaluation: Audio Understanding Error}
\end{figure*}
\newpage

\begin{figure*}[!htbp]
\centering
\includegraphics[width=\textwidth]{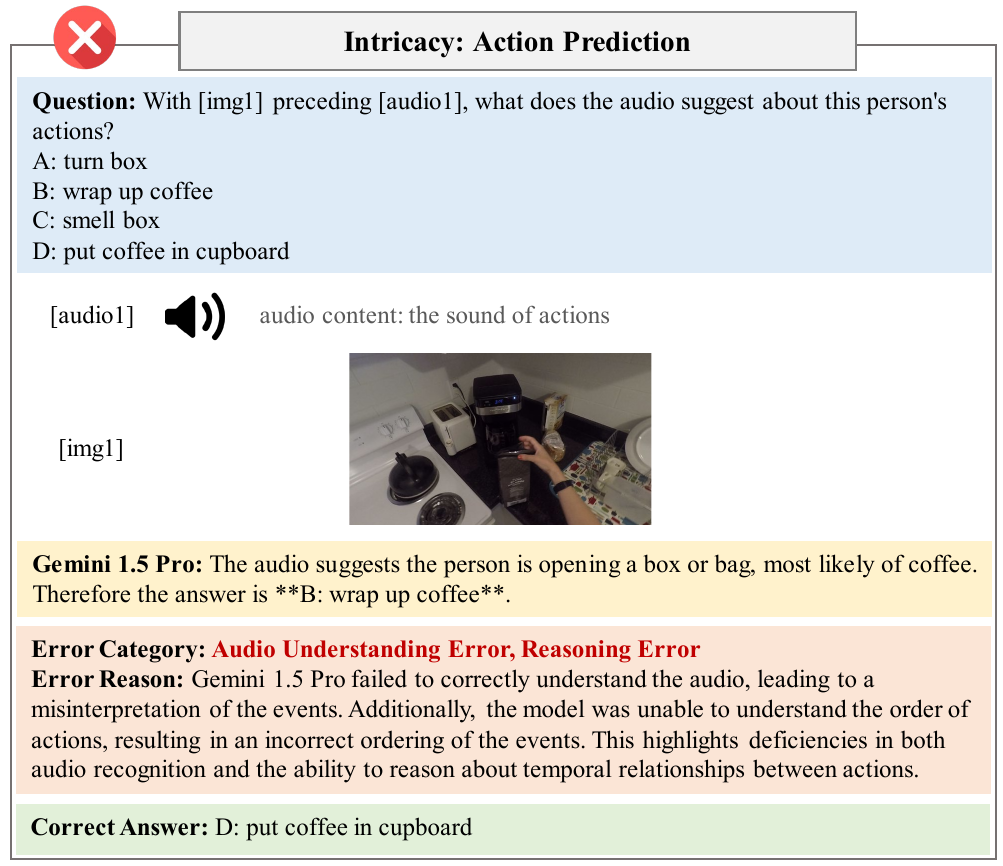}
\caption{A sampled error case in the action prediction task.}
\addcontentsline{afg}{appfigures}{\protect\numberline{\thefigure}Intricacy, Action Prediction: Audio Understanding Error, Reasoning Error}
\end{figure*}
\newpage

\begin{figure*}[!htbp]
\centering
\includegraphics[width=\textwidth]{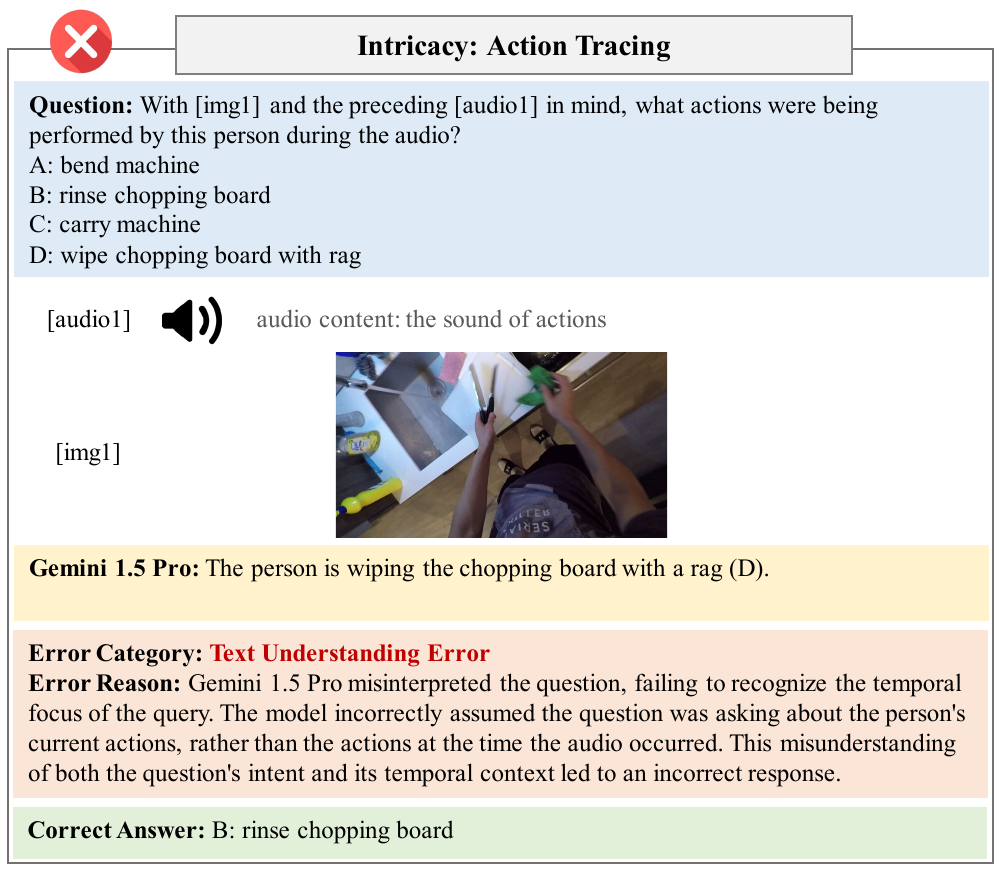}
\caption{A sampled error case in the action tracing task.}
\addcontentsline{afg}{appfigures}{\protect\numberline{\thefigure}Intricacy, Action Tracing: Text Understanding Error}
\end{figure*}
\newpage

\end{document}